\pdfoutput=1
\documentclass[11pt]{article}
\usepackage[preprint]{acl}

\usepackage{times}
\usepackage{latexsym}
\usepackage[T1]{fontenc}
\usepackage{microtype}
\usepackage{inconsolata}
\usepackage{graphicx}
\usepackage{amsmath}
\usepackage{algorithm}
\usepackage{algorithmicx}
\usepackage{algpseudocode}
\usepackage{graphicx}
\usepackage{multirow}
\usepackage{multicol}
\usepackage{enumitem}
\usepackage{booktabs}
\usepackage{tabularx}
\usepackage{todonotes}
\usepackage{adjustbox}
\usepackage{makecell}

\usepackage[most]{tcolorbox}
\usepackage{listings}

\usepackage{pgf}
\newcommand{\metric}[2]{%
  \begingroup
  \def\curr{#1}%
  \def\base{#2}%
  \count0=\numexpr#1-#2\relax
  \ifnum\count0>0
    \def\arrow{$\uparrow$}\def\clr{green!60!black}%
  \else\ifnum\count0<0
    \def\arrow{$\downarrow$}\def\clr{red!60!black}%
  \else
    \def\arrow{=}\def\clr{black}%
  \fi\fi
  \count2=\count0
  \ifnum\count2<0
    \multiply\count2 by -1
  \fi
  \curr\textcolor{\clr}{~(\arrow~\the\count2)}%
  \endgroup
}

\newcommand{\llama}[0]{LLaMA}
\newcommand{\llamas}[0]{LLaMA }
\newcommand{\olmo}[0]{OLMo}
\newcommand{\olmos}[0]{OLMo }
\newcommand{\repo}{\url{https://github.com/sislab-unitn/Knowledge-Dynamics-LLMs}}

\title{What Does Loss Optimization Actually Teach, If Anything? \\ Knowledge Dynamics in Continual Pre-training of LLMs}

\author{Seyed Mahed Mousavi\textsuperscript{*}, Simone Alghisi\thanks{Equal Contribution}, Giuseppe Riccardi\\
      Signals and Interactive Systems Lab, University of Trento, Italy \\
        \texttt{ \{mahed.mousavi,s.alghisi,giuseppe.riccardi\}@unitn.it}}
\begin{document}
\maketitle
\begin{abstract}
Continual Pre-Training (CPT) is widely used for acquiring and updating factual knowledge in LLMs. This practice treats loss as a proxy for knowledge learning, while offering no grounding into how it changes during training. We study CPT as a knowledge learning process rather than a solely optimization problem. We construct a controlled, distribution-matched benchmark of factual documents and interleave diagnostic probes directly into the CPT loop, enabling epoch-level measurement of knowledge acquisition dynamics and changes in Out-Of-Domain (OOD) general skills (e.g. math). We further analyze how CPT reshapes knowledge circuits during training. Across three instruction-tuned LLMs and multiple CPT strategies, optimization and learning systematically diverge as loss decreases monotonically while factual learning is unstable and non-monotonic. Acquired facts are rarely consolidated, learning is strongly conditioned on prior exposure, and OOD performance degrades from early epochs. Circuit analysis reveals rapid reconfiguration of knowledge pathways across epochs, providing an explanation for narrow acquisition windows and systematic forgetting. These results show that loss optimization is misaligned with learning progress in CPT and motivate evaluation of stopping criteria based on task-level learning dynamics.\footnote{Code and data available at \repo}
\end{abstract}

\section{Introduction}

\begin{figure*}[!ht]
    \centering
    \includegraphics[width=0.9\linewidth]{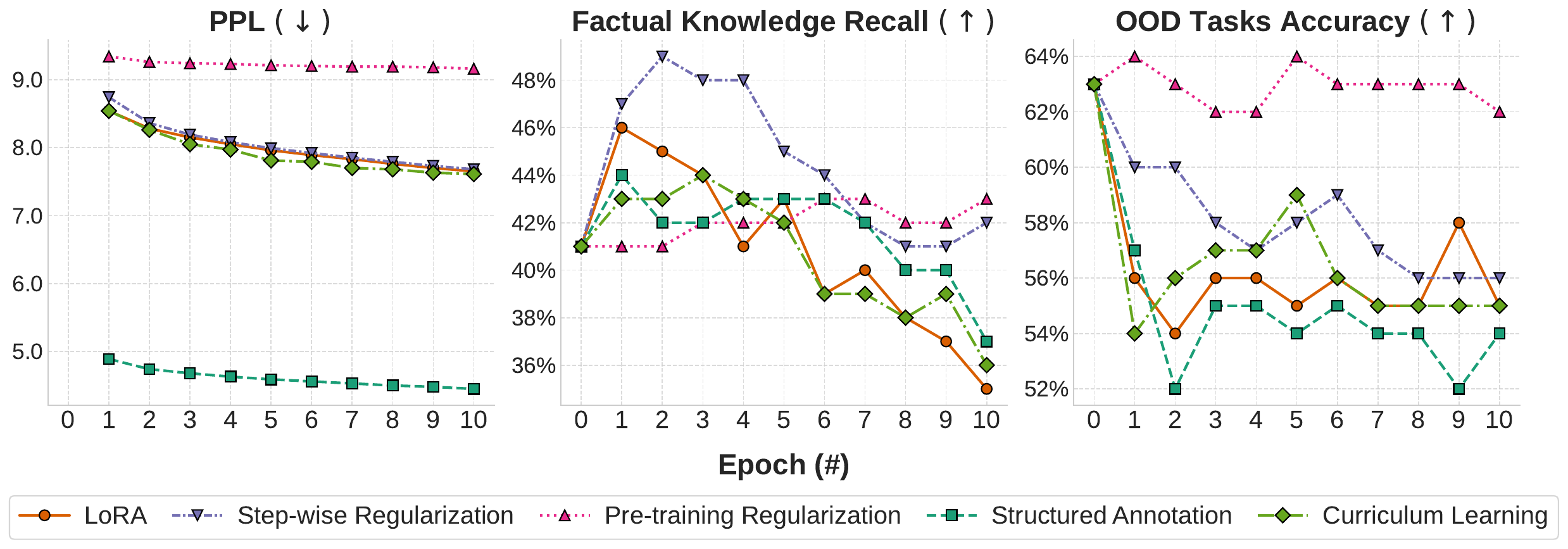}
    \caption{\textbf{Probing \olmo\textsubscript{7B} across continual pre-training epochs}. Across training strategies, perplexity (\textbf{PPL}) improves monotonically, as expected; however, knowledge acquisition is unstable, and \textbf{Factual Recall}'s gains and drops are erratically interleaved across epochs. Meanwhile, \textbf{OOD Tasks Accuracy} degrades monotonically through continued loss optimization.}
    \label{fig:olmo_loss-vs-learning}
\end{figure*}

Continual Pre-Training (CPT) is a standard approach for acquiring and updating unseen factual knowledge in Large Language Models (LLMs) by leveraging additional next-token prediction on revised corpora \cite{yin-etal-2025-improving}. In CPT, progress is assessed through loss reduction \cite{chen-etal-2025-towards-effective}, under the assumption that lower perplexity reflects successful acquisition or revision of factual knowledge. Training duration is typically chosen heuristically, and evaluation is performed only at convergence, leaving no mechanism for identifying when factual acquisition stabilizes or when continued optimization induces degradation.

Recent work suggests that this reliance on optimization signals during training leads to brittle, counterintuitive behavior. Predictions of training loss do not correlate with specific downstream tasks \cite{grattafiori2024llama}, small updates can degrade or destabilize performance \cite{springer2025overtrained}, and models often drift back toward pre-training behavior even after fitting new data \cite{ji-etal-2025-language-models}. Apparent improvements may reflect exploitation of familiar pre-training patterns rather than uptake of new content, especially when fine-tuning data resembles the model’s original training mixture \cite{li-etal-2024-formality,ou-etal-2025-llms}. As a result, factual updates may appear briefly and then vanish, overwritten by continued optimization \cite{springer2025overtrained}. Despite these observations, there is a lack of a framework for distinguishing learning from transient optimization, and measuring how knowledge is integrated during training.

We address this gap by introducing a probe-based framework that treats CPT as an observable learning process. Instead of inspecting only loss curves and end-of-training checkpoints, we insert task probes inside the training loop to measure, at each epoch, whether newly introduced facts are acquired, retained, forgotten, or distorted, and how these changes interact with out-of-domain abilities. This approach reveals the temporal structure of learning, i.e., what is gained early, what is lost later, and how optimization reshapes knowledge over time. To isolate learning dynamics, we construct a controlled benchmark of $\approx$10k up-to-date revisions of Wikipedia documents drawn from the model’s pre-training corpus. Using Dolma \cite{soldaini-etal-2024-dolma}, an open pre-training dataset, we replace documents seen during pre-training with their most recent revisions, preserving style and distribution while modifying only factual content. This design approximates a best-case scenario for knowledge acquisition under CPT, and allows observed failures to be attributed to learning dynamics rather than to domain shift, stylistic mismatch, or unknown data exposure. Building on this benchmark, we introduce two families of probes: knowledge probes that track factual knowledge acquisition during training, and out-of-domain probes, including social reasoning \cite{sap-etal-2019-social} and MMLU \cite{DBLP:journals/corr/abs-2009-03300}, that measure collateral effects on general skills.

We apply this framework to three instruction-tuned LLMs and commonly used strategies, LoRA \cite{hu2022lora}, structured annotation \cite{mousavi2025llms}, curriculum learning \cite{platanios-etal-2019-competence}, and KL-based regularization \cite{schulman2017proximal}. Across all models and methods, the following patterns emerge: a) \textit{Optimization diverges from learning:} perplexity decreases monotonically while knowledge acquisition is unstable and non-monotonic, and changes in acquisition, retention, and forgetting are not reflected in loss trends; b) \textit{Acquisition is strongly frequency-dependent:} improvements are driven mostly by high-frequency entities, while low-frequency entities show weak, short-lived, or no acquisition; c) \textit{Continued pre-training induces interference and forgetting:} degrading newly introduced, as well as prior knowledge and unrelated skills.

We contextualize these failures with three additional analyses. Retrieval-Augmented Generation (RAG) \cite{alghisi-etal-2024-fine-tune} establishes an upper bound on factual recall, showing that the relevant information can be recovered at inference time even when CPT fails to internalize it. Extending CPT to 100 epochs on a reduced, noise-free corpus amplifies forgetting rather than stabilizing learning. Finally, knowledge circuit analysis reveals rapid reconfiguration of internal subgraphs associated with factual recall across epochs, providing a structural explanation for transient acquisition and systematic overwriting.

Taken together, these findings indicate that, even under distribution-matched conditions, loss-driven optimization provides an unreliable signal for acquiring and consolidating factual knowledge, exposing structural limits in how LLMs integrate new information over the course of training. Our contributions are threefold:

\begin{itemize}[noitemsep,topsep=1pt,parsep=3pt,partopsep=3pt]
\item \textbf{A controlled evaluation setting} that leverages distribution-matched data to study knowledge acquisition without distributional confounds;
    \item \textbf{An epoch-level probing framework} that tracks knowledge acquisition and interference with general skills during CPT, complemented by circuit analysis that grounds these dynamics in changes to internal knowledge circuits;
    \item \textbf{A systematic comparison of CPT strategies across instruction-tuned LLMs}, highlighting fundamental misalignment between optimization objectives and knowledge acquisition.
\end{itemize}

\begin{figure*}[t!]
    \centering
    \includegraphics[width=0.65\linewidth]{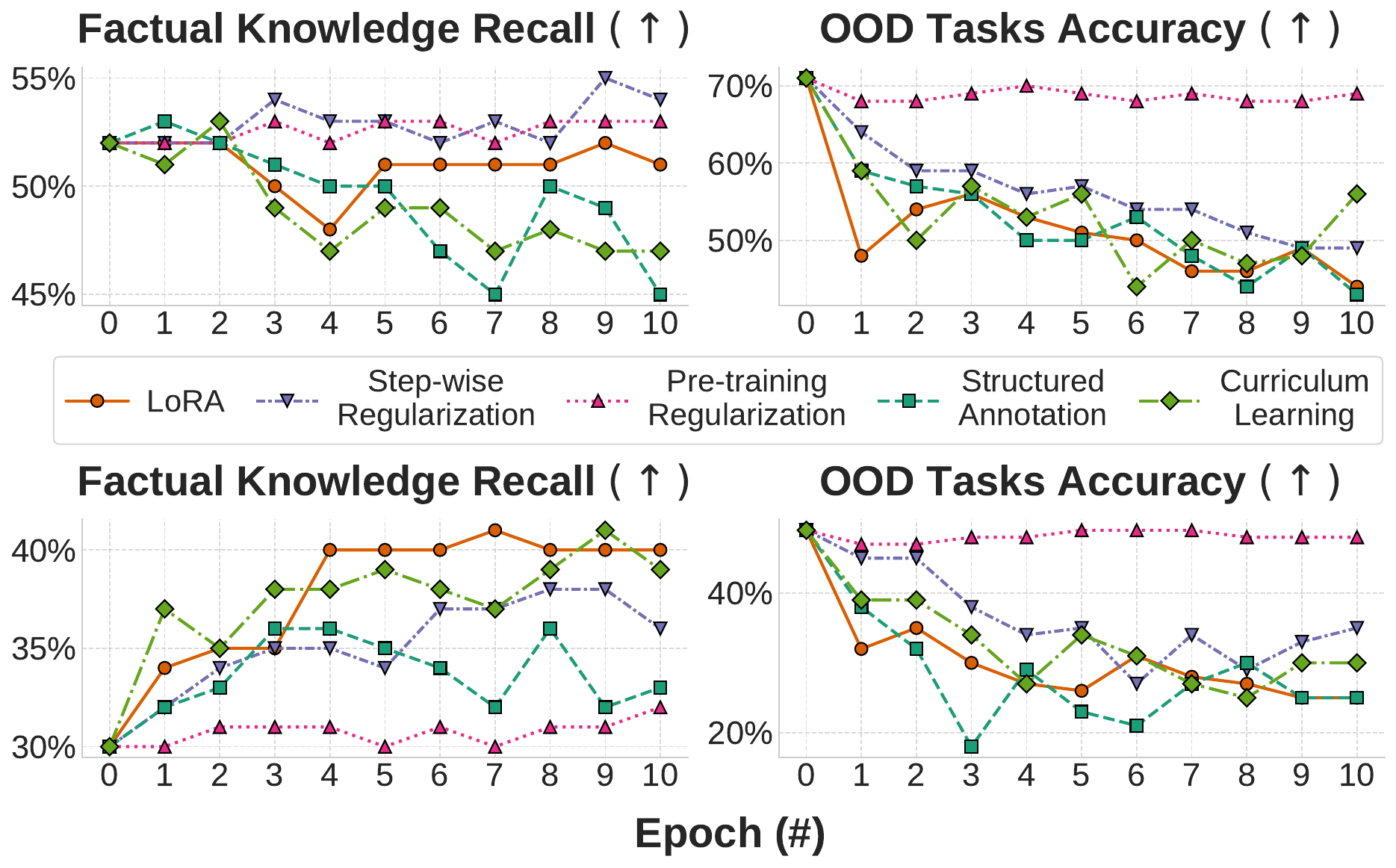}
    \caption{\textbf{Probing \llama\textsubscript{8B} (top) and \llama\textsubscript{1B} (bottom) across continual pre-training epochs} Across models, knowledge acquisition remain unstable. \textbf{Factual Recall} in \llama\textsubscript{8B} fluctuates around its pretrained baseline, while \llama\textsubscript{1B} exhibits transient gains followed by drops in the last epochs (7 and 9). \textbf{OOD Tasks Accuracy} drops immediately for all strategies except pre-training regularization.}
    \label{fig:llama_loss-vs-learning}
\end{figure*}

\section{Approach}

Let $M_0$ be a base model and $D$ a corpus containing factual documents. CPT $M_0$ on $D$ yields a sequence of checkpoints ${M}_{t=\{1,\ldots,T\}}$. Instead of evaluating only the final checkpoint, we treat this sequence as a learning trajectory. At each epoch, we measure whether $M_t$: i) incorporates the new factual information in $D$ (\emph{acquisition}), ii) preserves previously correct knowledge (\emph{retention}), and iii) maintains capabilities unrelated to the update (\emph{distortion}).

\subsection{Controlled Knowledge Corpus} Following evidence that LLMs learn more reliably from formal distribution-matched text \cite{li-etal-2024-formality}, we construct a corpus of up-to-date revisions of Wikipedia documents drawn directly from the model’s pre-training corpus. Using the Wikipedia portion of Dolma \citep{soldaini-etal-2024-dolma}, we select a total of 100 PERSON and GPE entities, spanning high- and low-frequency groups. For each entity, we retrieve its main Wikipedia page and the top 99 supporting pages containing the most mentions, resulting in a corpus of 9{,}171 documents (after removing duplicates). We then replace all documents with their most recent Wikipedia revisions, which introduce unseen factual content while preserving linguistic structure. This setting allows us to attribute model behavior directly to learning dynamics rather than distribution shift. Additional details are reported in Sections §\ref{subsec:appendix-corpus-construction} and §\ref{subsec:appendix-dataset}.

\subsection{Epoch-Level Probing Framework}
To observe learning as it unfolds, we evaluate each model using two families of probes. Probes are diagnostic only and play no role in optimization. All probes use zero-shot, instruction-style prompting and remain fixed across epochs, ensuring that changes reflect training dynamics rather than evaluation drift.

\paragraph{I. Knowledge Probes.} 
For each entity, we construct a set of ten manually curated probes derived from its main Wikipedia page. These probes target five complementary aspects of factual behavior, including time-sensitivity \cite{mousavi-etal-2024-dyknow}, factual recall, temporal understanding, entity linking, and output consistency. Responses are scored using ROUGE-1 recall against the ground truth answer. For each entity, we report the mean recall across its ten probe questions. Full probe construction details appear in Section §\ref{subsec:appendix-know-probes}.

\paragraph{II. Out-of-Domain Probes.}
To quantify interference with unrelated skills, we evaluate: a) 100 validated SocialIQA reasoning questions \cite{sap-etal-2019-social}, b) the full 1,400-question MMLU set \cite{DBLP:journals/corr/abs-2009-03300}. Performance is measured using exact-match accuracy, following standard multiple-choice QA evaluation protocols. We report the prompts used to answer multiple-choice questions in §Figure \ref{fig:mmlu-prompt}.

\begin{figure*}[!ht]
    \centering
    \includegraphics[width=\linewidth]{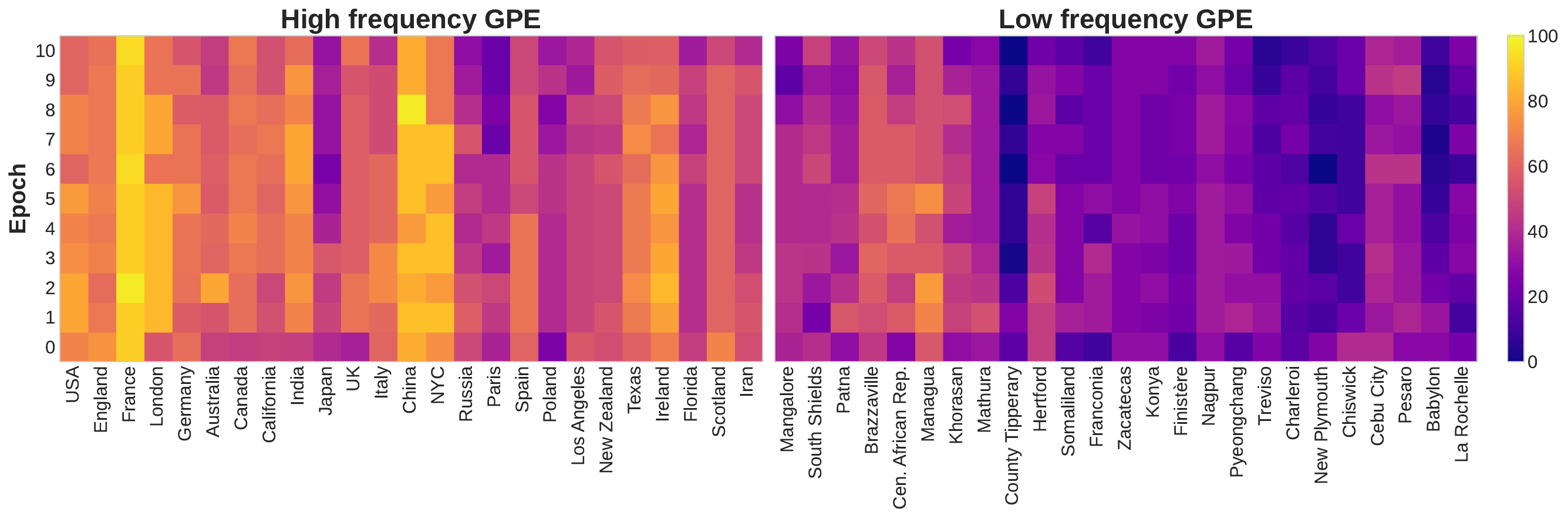}
\caption{\textbf{Named Entity (GPE) learning dynamics for high- and low-frequency entities via LoRA across continual pre-training epochs (\olmo)}. Low-frequency entities show minimal learning, while high-frequency entities exhibit non-monotonicity across epochs, alternating between learning ($\text{recall} \geq 60\%$) and forgetting ($\text{recall} < 60\%$). Results of other models and entity categories are shown in §Figures \ref{fig:olmo_lora_freq_heat_person}--\ref{fig:llama3-8b_lora_freq_heat_person}).}

    \label{fig:olmo_lora_freq_heat}
\end{figure*}

\paragraph{Epoch-Level Evaluation Cycle.}
At the end of each epoch $t$, we evaluate $M_t$ on both the knowledge and the OOD probes. Let $r_{e,t}$ be the average recall of entity $e$ at epoch $t$. We define \textit{acquisition} as $r_{e,t-1} < 60\%$ and $r_{e,t} \geq 60\%$; \textit{retention} when $r_{e,t-1} \geq 60\%$ and $r_{e,t} \geq 60\%$; and,  \textit{forgetting} when $r_{e,t-1} \geq 60\%$ but $r_{e,t} < 60\%$. The 60\% threshold corresponds to correct responses on a strict majority of probes (6 out of 10), balancing robustness with sensitivity to emerging learning signals. For out-of-domain tasks, we define \textit{distortion} as any decrease in out-of-domain performance relative to epoch $t-1$. This cycle yields a trajectory-level view of when knowledge is gained, consolidated, or lost, and when updates interfere with broader abilities.

\subsection{CPT Strategies} 
We evaluate different strategies commonly used in CPT, including data engineering, adaptive exposure, and explicit drift constraints. All methods employ parameter-efficient updates using LoRA. Additional implementation details are provided in Section §\ref{subsec:appendix-models}.

\paragraph{1) LoRA.} \cite{hu2022lora} We apply standard LoRA, which serves as the base technique and reflects a common practice for factual knowledge acquisition in LLMs. 

\paragraph{2) Structured Annotation}\cite{mousavi2025llms}. We augment the training corpus with lightweight entity tags of the form \texttt{<TAG> entity </TAG>} (Section §\ref{subsec:appendix-dataset}), to help the model associate lexical variants of an entity with a stable underlying reference, encouraging robust entity representations.

\paragraph{3) Curriculum Learning} \cite{platanios-etal-2019-competence}. For each epoch $t$, we construct $D_t$ by excluding entities (and corresponding documents) whose factual knowledge is acquired in the previous epoch ($r_{e,t-1}\ge 60\%$); while keeping (or, bringing back) the entities and corresponding documents with $r_{e,t-1} < 60\%$. This concentrates optimization on entities whose factual knowledge has not yet been learned and reduces over-exposure to already-learned material.

\paragraph{4) Divergence Regularization}\cite{schulman2017proximal}. To mitigate drift, we experiment with augmenting the loss with a divergence penalty: $\mathcal{L}_t = \mathcal{L}_{\mathrm{CE}}(M_t, D_t) + \lambda\,\mathrm{KL}(M_{\mathrm{ref}} \,\|\, M_t)$, where divergence is computed on logits of the updated documents. We evaluate (i) \textbf{\emph{pre-training regularization}} ($M_{\mathrm{ref}}=M_0$), constraining deviation from pretrained behavior; and (ii) \textbf{\emph{stepwise regularization}} ($M_{\mathrm{ref}} = M_{t-1}$), limiting abrupt changes across epochs. We use $\lambda = 10$ for (i) and $\lambda = 1$ for (ii), as they yield the best performance.

\subsection{Models} 

We evaluate \olmos Instruct 7B \cite{groeneveld-etal-2024-olmo}, \llamas 3.2 Instruct (1B), and 3.1 Instruct (8B) \cite{touvron2023llama}. \olmos is pretrained on Dolma, providing near-perfect distributional alignment with our documents. \llamas models differ in scale and pre-training mixture; although their training corpora are not public, recent studies indicate substantial overlap between modern LLM pre-training data \cite{soldaini-etal-2024-dolma}. This setup enables testing whether observed learning dynamics generalize across models and data regimes.

\section{Learning Dynamics with Loss Optimization}
We trace how factual knowledge and OOD abilities evolve under loss minimization epoch-by-epoch, and whether the optimization signal aligns with the intended task performance. We organize our findings around four central questions.

\subsection*{RQ1: How Well Do Models Acquire \& Retain Factual Knowledge?}
Across all models and strategies, the divergence between optimization and learning appears immediately. Knowledge acquisition is consistently non-monotonic, with early gains that do not accumulate and often reverse despite steady improvement in perplexity. In \olmo, the results presented in Figure \ref{fig:olmo_loss-vs-learning} indicate that LoRA and step-wise regularization produce sharp early improvements (reaching up to 49\% at epoch 2) that decline within a few epochs; structured annotation and curriculum learning briefly rise before declining; and pre-training regularization remains essentially flat (detailed results are presented in §Table \ref{tab:olmo_early_epochs_deltaood}). Regarding \llama\textsubscript{8B} and \llama\textsubscript{1B}, Figure~\ref{fig:llama_loss-vs-learning} shows the same pattern, including transient gains, followed by declines, despite consistent loss improvement (detailed results are presented in §Tables \ref{tab:llama1b_early_epochs_deltaood} and \ref{tab:llama8b_early_epochs_deltaood}). Crucially, \textbf{loss provides no indication of when learning succeeds or fails}. Peaks in factual accuracy do not align with local minima in perplexity, and continued optimization reliably pushes the model past its acquisition window into degradation. Probe-based trajectories reveal clear but narrow windows of successful acquisition invisible in the optimization signal.

\subsection*{RQ2: What Are Entity-Level Dynamics of Acquisition, Retention, \& Forgetting?} Entity-level trajectories presented in Figure \ref{fig:olmo_lora_freq_heat} ( and §Tables \ref{tab:olmo_early_epochs_deltaood}, \ref{tab:llama1b_early_epochs_deltaood}, \ref{tab:llama8b_early_epochs_deltaood} for all models) show that entities are repeatedly acquired and then forgotten across epochs, and no strategy yields monotonic learning. LoRA produces the largest fluctuations with rapid early acquisition followed by equally rapid loss, with forgetting increasing in later epochs (e.g., \olmos forgetting up to 7 entities; while \llama\textsubscript{1B} repeatedly forgets 5/6 entities). Structured annotation behaves similarly, with modest initial gains but accelerated degradation, particularly for low-frequency entities. Curriculum learning improves early acquisition for high-frequency entities yet does not stabilize retention. Step-wise regularization offers the most stable behavior, retaining the largest sets of entities ($\approx$28 in \olmo) and reducing forgetting relative to other methods, but it remains non-monotonic and continues to lose newly acquired items. Pre-training regularization minimizes drift but suppresses learning, yielding almost no acquisition. The repeated acquisition-forgetting cycles indicate that \textbf{acquired knowledge is not consolidated but is continually overwritten as optimization proceeds}. These dynamics explain the non-monotonic aggregate behavior observed in RQ1, i.e., CPT does not gradually integrate new knowledge, but instead produces unstable oscillations driven by overwriting the learned knowledge.

\begin{figure*}[t!]
    \centering
    \includegraphics[width=0.9\linewidth]{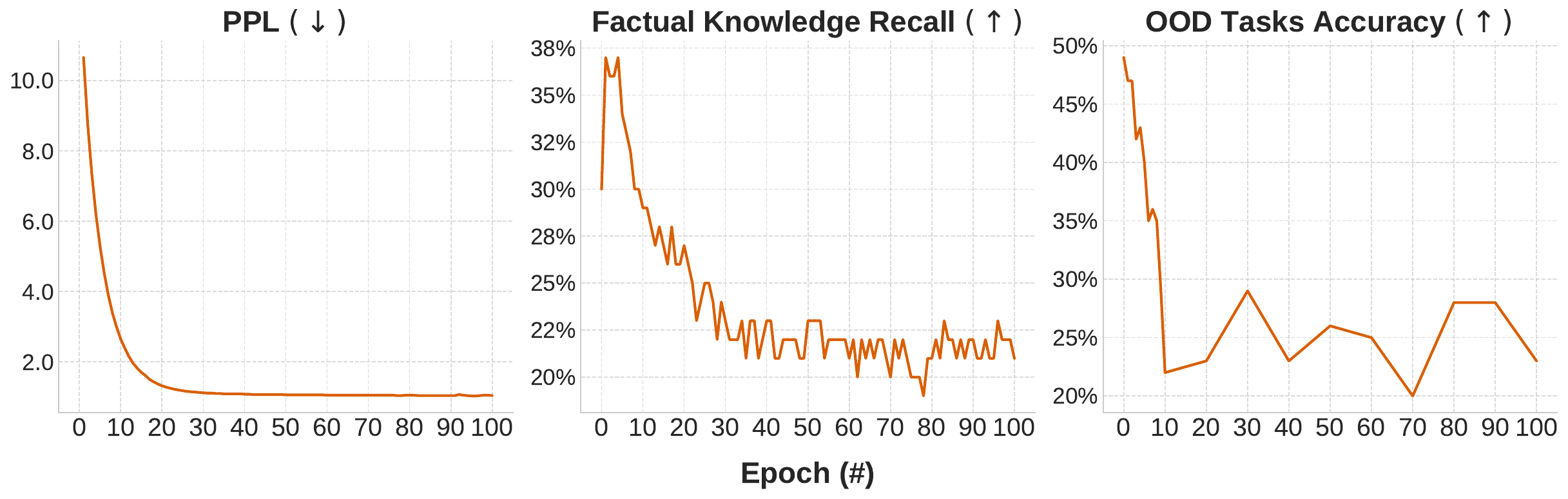}
       \caption{\textbf{Extended continual pre-training of \llama\textsubscript{1B} on the main entity documents for 100 epochs via LoRA.} \textbf{PPL} decreases monotonically and plateaus early, while \textbf{Factual Recall} peaks at epoch~3–5 and then collapses to a low, oscillatory steady state. \textbf{OOD Tasks Accuracy} drops sharply within the first 5–10 epochs and does not recover.}
\label{fig:extended_ft}
\end{figure*}

\subsection*{RQ3: How Does Pre-Training Frequency Shape the Learnability of Knowledge?}
Entity frequency is the strongest predictor of whether updated knowledge is ever acquired or retained. Figure \ref{fig:olmo_lora_freq_heat} presents the entity-level dynamics for GPE across CPT epoch on \olmos via LoRA (results of other models and entity categories are shown in §Figures \ref{fig:olmo_lora_freq_heat_person}--\ref{fig:llama3-8b_lora_freq_heat_person})

\begin{table}[t]
\centering
\begin{adjustbox}{max width=\linewidth}
\begin{tabular}{lllll}
\toprule
\multirow{2}{*}{\textbf{Model}} & \multirow{2}{*}{\textbf{Method}} &
\multicolumn{2}{c}{\textbf{\%Knowledge ($\Delta$)}} &
\multirow{2}{*}{{\textbf{\shortstack{\# Covered\\Entities ($\Delta$)}}}} \\
\cmidrule(l{4pt}r{4pt}){3-4}

 & &
\textbf{HighFreq}. & \textbf{LowFreq}. & 
\\
\midrule

\multirow{6}{*}{\textbf{\olmo}}

& Best & \metric{56}{49} & \metric{42}{33} & \metric{33}{17} \\
\cmidrule(l{4pt}r{4pt}){2-5}

& RAG \\

{} & \hspace{0.2cm} \textit{All}
& \metric{56}{49} & \metric{47}{33} & \metric{25}{17} \\

{} & \hspace{0.2cm}\textit{Main}
& \metric{62}{49} & \metric{57}{33} & \metric{36}{17} \\

{} & \hspace{0.2cm}\textit{Gold}
& \metric{64}{49} & \metric{61}{33} & \metric{42}{17} \\
\midrule

\multirow{6}{*}{\textbf{\llama\textsubscript{8B}}}

& Best & \metric{63}{60} & \metric{47}{44} & \metric{43}{36} \\
\cmidrule(l{4pt}r{4pt}){2-5}

& RAG \\

{} & \hspace{0.2cm} \textit{All}
& \metric{63}{60} & \metric{55}{44} & \metric{51}{36} \\

{} & \hspace{0.2cm}\textit{Main}
& \metric{66}{60} & \metric{65}{44} & \metric{65}{36} \\

{} & \hspace{0.2cm}\textit{Gold}
& \metric{67}{60} & \metric{70}{44} & \metric{73}{36} \\
\midrule

\multirow{6}{*}{\textbf{\llama\textsubscript{1B}}}

& Best & \metric{50}{38} & \metric{33}{22} & \metric{23}{12} \\
\cmidrule(l{4pt}r{4pt}){2-5}

& RAG \\

{} & \hspace{0.2cm} \textit{All}
& \metric{56}{38} & \metric{44}{22} & \metric{39}{12} \\

{} & \hspace{0.2cm}\textit{Main}
& \metric{59}{38} & \metric{54}{22} & \metric{48}{12} \\

{} & \hspace{0.2cm}\textit{Gold}
& \metric{62}{38} & \metric{58}{22} & \metric{52}{12} \\
\bottomrule
\end{tabular}
\end{adjustbox}

\vspace{-2mm}
\caption{\textbf{RAG as an upper bound on factual update performance.}  
We compare RAG to the \textit{Best} CPT checkpoint (presented in Figures \ref{fig:olmo_loss-vs-learning} and \ref{fig:llama_loss-vs-learning}, i.e. step-wise regularization at epoch~2 for \olmo, epoch~9 for \llama\textsubscript{8B}, and LoRA at epoch~7 for \llama\textsubscript{1B}). RAG is evaluated with retrieval from the pool of \textit{All} documents, \textit{Main} documents, \textit{Gold} document. Green deltas indicate improvements over the base-model performance.}
\label{tab:rag}
\end{table}

Across models, high-frequency entities exhibit consistent acquisition-forgetting cycles, i.e., performance improves briefly in early epochs and then degrades, with no monotonic progression toward stable retention. Low-frequency entities, by contrast, show little to no learning signal,  with gains that are small and short-lived; several entities are never acquired under any model. This asymmetry persists even in \olmo, where the training corpus is drawn directly from the pre-training distribution, ruling out domain mismatch as an explanation. Instead, these results indicate that \textbf{factual acquisition is strongly conditioned on prior exposure}. High-frequency entities receive sufficient contextual signal to be learned transiently, while low-frequency entities lack the representational support needed for meaningful or durable updates.

\textbf{RQ4: How Does Continual Pre-Training Interact With Unrelated Skills?}

Across models and strategies, CPT consistently disrupts out-of-domain abilities (Figures \ref{fig:olmo_loss-vs-learning}, and \ref{fig:llama_loss-vs-learning}). As anticipated, pre-training regularization preserves OOD performance by constraining drift; consequently, it thereby suppresses learning. Regarding other strategies, we observe an immediate and persistent decline in OOD performance in the very first epoch, well before optimization converges and even in epochs where factual accuracy temporarily improves. These dynamics reveal a trade-off, i.e., \textbf{strategies that permit factual updates also induce interference, and strategies that prevent interference also prevent learning}. No method achieves both stable acquisition and stable general capabilities.

\section{Upper Bound \& Circuit Analysis}
Previous section's results reveal fundamental instability in how loss optimization supports knowledge acquisition during training. To contextualize these limitations, we examine a) attainable upper-bound performance using retrieval-augmented generation (RAG) \cite{alghisi-etal-2024-fine-tune}, b) the behavior of models trained beyond standard horizons (100 epochs), and c) the impact of CPT on internal knowledge circuits. Together, these analyses clarify what models \emph{could} achieve under idealized conditions, how optimization behaves when pushed to extremes, and what internal representations change, or fail to change, when CPT attempts to integrate novel factual knowledge.

\subsection{RAG: Upper Bound}
We evaluate RAG  as an upper bound on factual recall, supplying the revised documents directly at inference time. We compare RAG to the strongest CPT checkpoints per model, i.e., step-wise regularization at epoch~2 for \olmo, epoch~9 for \llama\textsubscript{8B}, and LoRA at epoch~7 for \llama\textsubscript{1B} (§Tables \ref{tab:olmo_early_epochs_deltaood}, \ref{tab:llama1b_early_epochs_deltaood}, and \ref{tab:llama8b_early_epochs_deltaood}). We consider three settings of decreasing complexity where the correct passage must be retrieved from the pool of 1) \textit{All} documents ($\approx$10k); 2) \textit{Main} documents (100); or 3) \textit{Gold} single document per query. These conditions separate errors due to retrieval from those due to the model’s own ability to use the provided evidence. Across all models, RAG consistently outperforms the best CPT epochs, Table \ref{tab:rag}. Even in the noisy \textit{`All'} setting, RAG matches or exceeds the gains achieved by parameter updates, showing that the new information is accessible at inference time but not reliably internalized through optimization. Restricting retrieval to the \textit{`Main'} and \textit{`Gold'} documents yields further gains, with each model reaching its effective ceiling, implying that most remaining errors arise from retrieval noise. Overall, these results show that the limitation lies in the optimization process, not in the training documents. RAG consistently recovers information that CPT either fails to acquire or subsequently forgets.

\begin{figure}[t]
    \centering
    \includegraphics[width=0.75\linewidth]{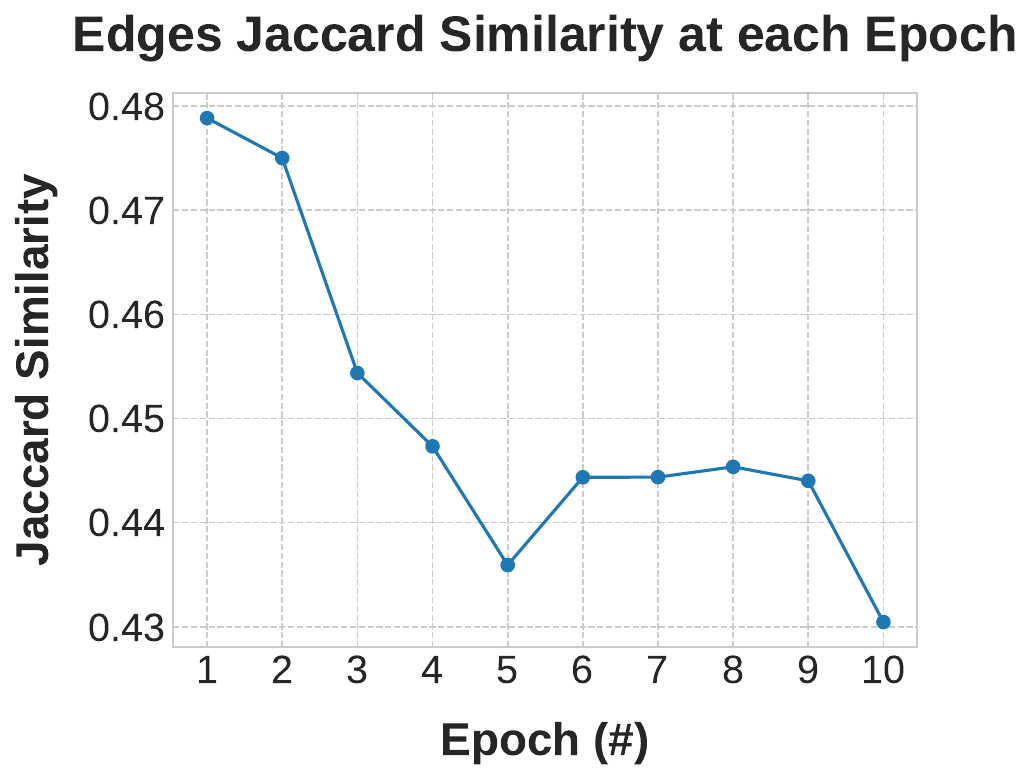}
\caption{\textbf{Jaccard similarity between the base model knowledge circuit $C_0$ and the circuits extracted at each epoch of CPT $C_t$ (\llama\textsubscript{1B})}. Circuit overlap drops sharply after the first epoch, with the majority of original edges being replaced immediately.}
\label{fig:jaccard}
\end{figure}

\subsection{Extended Continual Pre-Training: Does More Optimization Help?}
To test whether instability stems from insufficient training, we extend CPT on \llama\textsubscript{1B} using LoRA to 100 epochs on the 100 \textit{main} Wikipedia pages of the entities, removing noise and giving the model repeated, unambiguous exposure to the novel factual knowledge. The resulting trajectories shown in Figure \ref{fig:extended_ft} reveal that longer training does not stabilize learning. Perplexity decreases rapidly and plateaus, yet factual accuracy peaks within the first few epochs ($\approx$37-38\% at epochs 3-5) before collapsing into a persistent low-accuracy regime ($\approx$20-23\%). This decline continues despite perfectly aligned data, ruling out explanations based on retrieval noise, distributional mismatch, or insufficient exposure. The model simply fails to consolidate the updated facts, even after 100 direct passes over them. Out-of-domain performance deteriorates just as sharply. Accuracy drops from ~50\% to below 30\% within the first 10 epochs and never recovers, despite perplexity convergence. Taken together, these results show that ``training longer'' does not bring the model closer to stable learning or to the retrieval-based upper bound. Extended CPT amplifies the core pathology, i.e., factual updates are acquired briefly, then overwritten, while unrelated capabilities degrade irreversibly. Rather than converging toward the revised knowledge, the model settles into a low-recall equilibrium that is disconnected from the optimization signal itself.

\subsection{Mechanistic Interpretability}
To explain the learning dynamics instability reported in previous sections, we analyze how CPT reshapes model’s \emph{knowledge circuits}. Following prior work \cite{ou-etal-2025-llms}, we identify knowledge circuits using the EAP-IG attribution method applied to factual triplets derived from high-frequency entities. Similar to the previous experiment, we focus on \llama\textsubscript{1B}, whose scale enables detailed circuit-level analysis, and perform five runs to account for variations. Further implementation details are provided in Appendix Section §\ref{subsec:appendix-mechanistic}.

\begin{figure}[t]
    \centering
    \includegraphics[width=\linewidth]{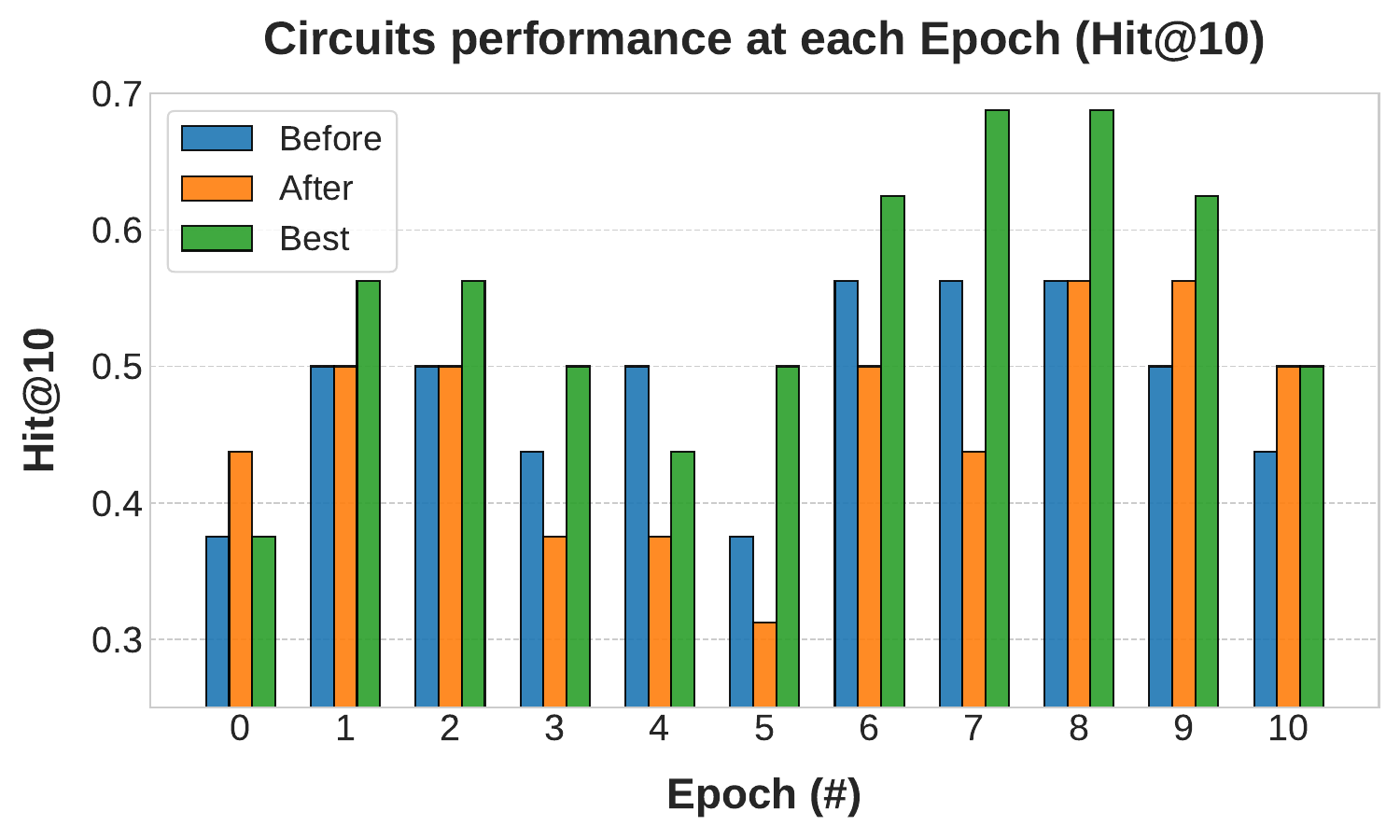}
\caption{\textbf{Circuit-level performance (Hit@10) at different CPT epochs (LoRA).} We compare the base-model circuit (\emph{Before}, $C_0$), the final checkpoint circuit (\emph{After}, $C_{10}$), and the circuit extracted at the epoch with the peak factual accuracy (\emph{Best}, $C_7$) in \llama\textsubscript{1B}. The performance indicates that CPT does not consolidate internal knowledge subgraphs.}
\label{fig:circuits}
\end{figure}

Figure~\ref{fig:jaccard} compares the base-model circuit $C_0$ with circuits extracted at each epoch $C_t$. Circuit similarity drops sharply as soon as training begins; various edges in $C_0$ are replaced within the first epoch, and overlap continues to decline thereafter. This rapid turnover occurs despite the update corpus being distribution-matched to pre-training, indicating that CPT does not selectively refine existing factual pathways. Instead, attribution mass is repeatedly redirected toward different edges, effectively rewriting the circuitry responsible for factual recall. 

To assess the functional consequences of this reorganization, we examine the circuit-level performance, i.e., the correct token being among the top 10 predictions (Hit@10). We evaluate three circuits: the base-model circuit (\emph{Before}, $C_0$), the final checkpoint (\emph{After}, $C_{10}$), and the circuit from the epoch with peak factual performance (\emph{Best}, LoRA $C_7$). Figure~\ref{fig:circuits} reveals that circuit-level performance oscillates across epochs, closely mirroring the non-monotonic behavioral trajectories observed earlier. The \emph{Before} and \emph{After} circuits reach their maximum performance within the last epochs, while the \emph{Best} circuit attains the peak at epoch 7 (in line with model performance) but subsequently degrades.

These results indicate that CPT does not consolidate updated facts by strengthening a stable set of edges. Instead, it repeatedly reconfigures the subgraphs supporting factual recall, yielding transient pathways that are overwritten as optimization proceeds. This continual circuit-level reorganization provides a mechanistic explanation for the narrow acquisition windows, fragile retention, and systematic forgetting observed throughout this study.

\section{Discussion}
This study shows that loss optimization is not a reliable indicator of learning in continual pre-training. Across models and strategies, factual knowledge is acquired partially, retained unreliably, and frequently overwritten as optimization continues. These dynamics persist under distribution-matched data, repeated exposure, and extended training, ruling out data quality or insufficient optimization as explanations. Instead, they reflect a structural misalignment between loss minimization and knowledge consolidation. Circuit analysis shows that CPT repeatedly reconfigures the subgraphs responsible for factual recall, replacing earlier circuits with new ones. This continual reorganization aligns with the observed acquisition–loss cycles and explains why repeated exposure fails to produce consolidation. Besides, the upper-bound achieved by RAG demonstrates that the limitation lies not in lack of evidence in the data, but in the dynamics of parameter updates, which fail to internalize new facts without destabilizing existing representations.

Probe-based trajectories reveal \emph{narrow learning windows} in which factual knowledge is briefly acquired before being lost by further optimization. These windows do not widen with extended training and are invisible to loss curves, which continue to improve as factual accuracy peaks and then declines. Therefore, since loss does not offer an indication of acquisition or degradation, perplexity-based checkpoint selection systematically favors models in which updated knowledge has already decayed. That is, the checkpoint with the lowest loss is rarely the one with the most reliable knowledge. These results motivate a \emph{task-based stopping criterion}.  Probe-based signals reveal acquisition peaks that loss cannot detect, indicating that effective stopping must rely on task-level performance rather than next-token prediction.

Our findings call for a shift in how CPT is evaluated, away from last-epoch task performance and final checkpoints, and toward trajectory-level analyses that reveal when learning occurs, when it fails, and when continued optimization becomes harmful. More fundamentally, these results show that perplexity is a poor proxy for learning, as it provides no signal of acquisition, retention, or forgetting, and can continue to improve even as factual knowledge degrades.

\section{Literature Review}
CPT is commonly assumed to update LLM knowledge reliably, particularly when applied to well-edited, distribution-matched text \citep{li-etal-2024-formality}. Prior work shows that data engineering techniques \citep{chen-etal-2025-towards-effective} and structured representations \citep{mousavi2025llms} can further improve loss trends during CPT, yet these improvements provide limited insight into whether parameter updates correspond to stable changes in model knowledge. \citet{ou-etal-2025-llms} show that high-frequency information are learned more readily than rare or novel content. Other studies report that prolonged optimization can make models brittle in the absence of explicit regularization \citep{springer2025overtrained}, and often induces reversion toward pre-training behavior rather than consolidation of new information \citep{ji-etal-2025-language-models}. Recent analyses decompose finetuning dynamics at the level of gradient influence and token probabilities, revealing systematic interference effects during training \citep{renlearning}.

\section{Conclusion}
We investigated how loss optimization during continual pre-training relates to the acquisition and retention of factual knowledge in LLMs. We introduced a controlled distribution-matched benchmark, an epoch-level probing framework that tracks knowledge dynamics throughout training, and a systematic comparison of common training strategies under this lens. Together, these contributions provide a process-level view of CPT and decouples optimization progress from changes in model knowledge and capabilities. This perspective motivates future work on novel training objectives, evaluation protocols, and diagnostic tools that make learning dynamics observable during model adaptation rather than inferring them from loss alone.

\bibliographystyle{acl_natbib}
\bibliography{custom}

\clearpage

\appendix

\section{Appendix}
\label{sec:appendix}

\subsection{Knowledge Corpus Construction}
\label{subsec:appendix-corpus-construction}
\begin{figure}[ht!]
    \centering
    \includegraphics[width=\linewidth]{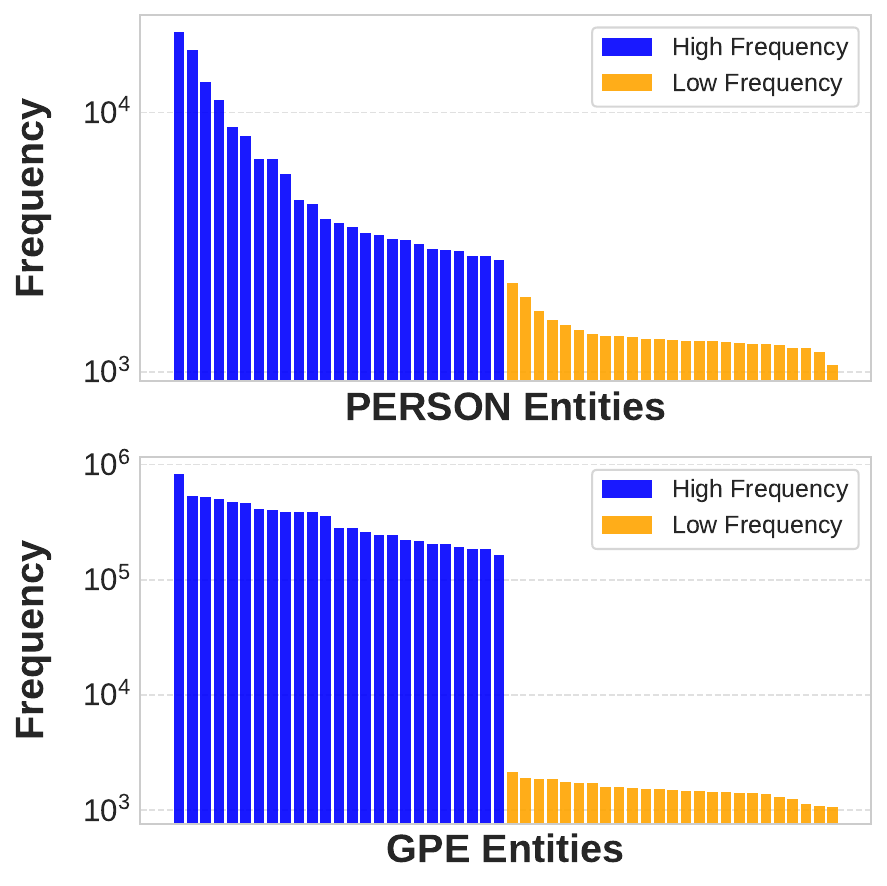}
\caption{\textbf{Entity frequency distribution for PERSON and GPE categories in the Dolma v1.5 corpus.}}
    \label{fig:entityfreq}
\end{figure}

We begin with the Wikipedia-Wikibooks partition of Dolma v1.5 \citep{soldaini-etal-2024-dolma}, comprising 6.14M documents (16GB). Using a named-entity recognizer (SpaCy\footnote{\url{https://spacy.io}} \texttt{en\_core\_web\_trf}), we extract PERSON and GPE entities and compute document-level frequencies across the partition. From each category, we sample 25 high-frequency and 25 low-frequency entities (100 total, reported in §Table \ref{tab:gpe-entities}). Low-frequency entities are required to appear in at least 1{,}000 documents in Dolma to ensure sufficient training signal. §Figure~\ref{fig:entityfreq} shows that the resulting groups are well-separated. Since some documents are shared among different entities, the final corpus contains 9{,}171 unique documents. Then, each document is replaced with its most recent version from Wikipedia (downloaded on January 2nd 2025).

\subsection{Training Dataset Construction}
\label{subsec:appendix-dataset}
We construct two versions of the training dataset for continual pre-training: a plain-text version, and a structured-annotation version in which named entity tags are added inline~\cite{mousavi2025llms}. 
For structured-annotation, only entities of type PERSON, NORP, ORG, and GPE are tagged. Entity tags are inserted inline using explicit start and end markers that surround each entity span (i.e., \texttt{<TAG> entity </TAG>}, where \texttt{TAG} is replaced with PERSON, NORP, ORG, or GPE depending on the entity).

Documents are tokenized using the \olmo's tokenizer (since it has the smallest context window among the models) and segmented into contiguous training samples of at most 2048 tokens. Segmentation respects entity boundaries: once an entity span is started, it is fully included before truncation is applied. This ensures that samples are identical in terms of content (i.e., factual knowledge), and only differ for the presence of entity tags.

\subsection{Knowledge Probe Construction}
\label{subsec:appendix-know-probes}
For each of the 100 selected entities, we prompt GPT-4o (\texttt{gpt-4o-2024-08-06}) to generate twenty candidate questions conditioned on the entity’s main Wikipedia page. 
Questions are designed to probe five facets of factual behavior:
\begin{itemize}[noitemsep,topsep=1pt,parsep=3pt,partopsep=3pt]
    \item \textbf{Time-sensitive knowledge} \citep{mousavi-etal-2024-dyknow}: 
    Questions about the entity’s current state, with answers that are valid within a specific time window and may change over time (e.g., such as roles, affiliations, or statuses).
    \item \textbf{Factual recall}: 
    Objective, document-grounded questions with precise, verifiable answers stated in the text.

    \item \textbf{Temporal understanding}: 
    Questions that explicitly reference a specific time period or temporal relation (e.g., before, after, during), requiring the model to interpret temporally scoped statements in the document rather than retrieving atemporal facts.

    \item \textbf{Entity linking / relational reasoning}: 
    Questions that explicitly connect the target entity to another entity mentioned in the document (e.g., a person, organization, or location), probing whether the model correctly identifies and binds relationships stated in the text.

    \item \textbf{Consistency}: 
    Semantically equivalent questions derived from the same underlying factual statement. Because both questions are answerable from the same document context, disagreement reflects inconsistency in representation rather than missing information.
\end{itemize}

We report the prompt used to generate the questions in §Figures \ref{fig:entity-prompt} and \ref{fig:consistency-prompt}. We manually review and filter these into a final set of ten probes per entity (i.e., 1{,}000), ensuring coverage, correctness, and linguistic diversity while avoiding answer leakage across probe categories.

\begin{figure*}[!ht]
    \begin{tcolorbox}[
        title={Prompt template for Entity-Specific Question Generation},
        colback=blue!5,
        colframe=blue!60,
        fonttitle=\bfseries,
    ]
\small
\begin{verbatim}
You are a helpful assistant for generating document-grounded questions.

Document text: {html_text}

The main entity in the document is "{entity_name}". Based on the document, generate 4 questions about
"{entity_name}" for each of the following categories, with short answers (max. 5 words):

**Factual Question**: Objective, precise questions with verifiable answers found in  the document.

**Time-Sensitive Question**: Questions about the entity's current state with no time specifier, 
with answers that are valid within a specific timeframe and change over  time (avoid numbers, 
dates, and percentages).

**Temporal Understanding Question**: Questions about events related to the entity, explicitly 
referencing a specific time or period (avoid asking When).

**Entity Linking Question**: Questions that connect "{entity_name}" with another entity in the text
(e.g., persons, locations, or organizations).

Always mention explicitly "{entity_name}" in the questions (avoid using pronouns).
Do not ask questions whose answers are dates, numbers, or statistics.
Do not ask questions that have multiple correct answers (avoid lists).

**Output Format**:
"{entity_name}"|[QuestionType]|[Question]|[Answer]|[Relevant Text Chunk Containing the Answer]

Ensure all questions are answerable using the provided document text.
        \end{verbatim}  
    \end{tcolorbox}
    \caption{Prompt template used to generate entity-specific probing questions from each entity's main Wikipedia document.}
    \label{fig:entity-prompt}
\end{figure*}

\begin{figure*}[!ht]
    \centering
    \begin{tcolorbox}[
        title={Prompt template for Consistency Question Generation},
        colback=blue!5,
        colframe=blue!60,
        fonttitle=\bfseries
    ]
\small
\begin{verbatim}
You are a helpful assistant for generating question rephrases.
Given a question, provide 3 different paraphrases and put them in a comma separated list
(i.e., [“rephrase1", “rephrase2", “rephrase3"]):

Question: {question}

Paraphrases:
\end{verbatim}
    \end{tcolorbox}
    \caption{Prompt template used to generate consistency probes by producing multiple semantically equivalent paraphrases of a given question.}
    \label{fig:consistency-prompt}
\end{figure*}

\begin{figure*}[!ht]
    \centering
    \begin{tcolorbox}[
        title={Prompt Template for OOD Task Evaluation},
        colback=blue!5,
        colframe=blue!60,
        fonttitle=\bfseries,
    ]
        \small
\begin{verbatim}
Be brief. Do not add any explanation. Only answer with 'A', 'B', 'C', or 'D'.

Q: {question}

A. {choice_A}
B. {choice_B}
C. {choice_C}
D. {choice_D}
\end{verbatim}
    \end{tcolorbox}
    \caption{Prompt used for evaluating performance on the MMLU and SocialIQA benchmark. The model is presented with a multiple-choice question and instructed to respond with a single option label without any additional explanation.}
    \label{fig:mmlu-prompt}
\end{figure*}

\subsection{Experimental details}
\label{subsec:appendix-models}
\textbf{Models} For our experiments, we used the following model checkpoints available from HuggingFace:
\begin{enumerate}[noitemsep,topsep=1pt,parsep=3pt,partopsep=3pt]
    \normalsize \item \olmos 7B Instruct, \small\url{https://huggingface.co/allenai/OLMo-7B-Instruct-hf}
    \normalsize \item \llamas 3.1 8B Instruct, \small\url{https://huggingface.co/meta-llama/Llama-3.1-8B-Instruct}
    \normalsize \item \llamas 3.2 1B Instruct, \small\url{https://huggingface.co/meta-llama/Llama-3.2-1B-Instruct}
\end{enumerate}

\textbf{CPT Parameters} We train each model for 10 epochs using LoRA (with scaling $\alpha = 32$ and rank $r = 64$), updating only the query and value matrices. All models are loaded in FP16 and optimized with AdamW. Based on preliminary experiments, we select a learning rate of $1 \times 10^{-4}$ for both \llamas models (1B and 8B) and $1 \times 10^{-5}$ for \olmo, as these settings yield lower perplexity. As in the pre-training stage, we present the data without a chat template.
Most experiments are run on a single A100 GPU (80 GiB) with a batch size of two, using a maximum sequence length of 2048 tokens. Each run requires at most 53 hours. Experiments involving KL regularization instead require two A100 GPUs (80 GiB each) and take up to 83 hours. 

\textbf{Generation} We generate all probe responses on a single NVIDIA GeForce RTX 3090 GPU with 24 GiB of memory, using a batch size of 4. For all experimental settings, responses are generated using greedy decoding, ensuring deterministic outputs and eliminating variability due to sampling. Generation is capped at 25 new tokens. 

\textbf{Evaluation} We evaluate knowledge probes using ROUGE-1\footnote{\url{https://github.com/google-research/google-research/tree/master/rouge}} recall. Out-of-domain (OOD) probes are derived from multiple-choice benchmarks and are evaluated using exact match\footnote{\url{https://huggingface.co/spaces/evaluate-metric/exact_match}}. For these benchmarks, we identify the predicted answer by applying a simple regular expression that extracts the first occurrence of \texttt{ABCD} for MMLU and \texttt{ABC} for SocialIQA.

\subsection{Mechanistic Interpretability}
\label{subsec:appendix-mechanistic}
We analyze the model’s knowledge circuits using the Edge Attribution Patching-Integrated Gradients (EAP-IG) method\footnote{\url{https://github.com/hannamw/eap-ig}}. Attributions are computed over attention heads and MLP layers using 30 integration steps based on the difference between clean and corrupted attribute logits. Based on previous results~\cite{ou-etal-2025-llms}, we retain the 50k edges with the highest absolute attribution magnitude; these edges define the directed computational graph that we refer to as a knowledge circuit. Factual knowledge triplets are converted into natural-language inputs using a fixed template, \texttt{``The name of the current head of state of \{\} is''}, where the placeholder is filled with countries drawn from our set of high-frequency entities. The full list of subjects and their corresponding cleaned gold head-of-state labels is shown in Table~\ref{tab:heads-of-state-triplets}. To construct counterfactual inputs, we generate a corrupted version of each query by replacing the original subject with a random country from the same set. Knowledge circuits are then identified by computing the logit difference between model completions for the clean and corrupted inputs. To quantify changes in circuit structure induced by CPT, we compare circuits before and after training using the Hit@10 metric~\cite{ou-etal-2025-llms}.

\begin{table}[!ht]
  \centering
  \small
  \begin{tabular}{ll}
    \toprule
    \textbf{Subject} & \textbf{Head of state (gold label)} \\
    \midrule
    \textit{USA}      & \textit{Donald Trump} \\
    \textit{England}  & \textit{Charles III} \\
    \textit{France}   & \textit{Emmanuel Macron} \\
    \textit{Germany}  & \textit{Frank-Walter Steinmeier} \\
    \textit{Austria}  & \textit{Alexander Van der Bellen} \\
    \textit{Canada}   & \textit{Charles III} \\
    \textit{India}    & \textit{Droupadi Murmu} \\
    \textit{Japan}    & \textit{Naruhito} \\
    \textit{Italy}    & \textit{Sergio Mattarella} \\
    \textit{China}    & \textit{Xi Jinping} \\
    \textit{Russia}   & \textit{Vladimir Putin} \\
    \textit{Spain}    & \textit{Felipe VI of Spain} \\
    \textit{Poland}   & \textit{Karol Nawrocki} \\
    \textit{Ireland}  & \textit{Catherine Connolly} \\
    \textit{Scotland} & \textit{Charles III} \\
    \textit{Iran}     & \textit{Masoud Pezeshkian} \\
    \bottomrule
  \end{tabular}
  \caption{Subjects and corresponding gold head-of-state labels used to construct factual knowledge triplets.}
  \label{tab:heads-of-state-triplets}
\end{table}

We run all circuit-extraction and evaluation experiment on a single NVIDIA GeForce RTX 3090 GPU with 24 GiB of memory. To obtain more reliable results, we perform a total of five runs and save the circuit with the best Hit@10 performance. Notably, we observe no variation among our experiments in Hit@10 and a negligible difference between clean and corrupted logits (i.e., standard deviation $1 \times 10^{-7}$).

\begin{table*}[ht!]
\centering
\small
\begin{adjustbox}{max width=\textwidth}
\begin{tabular}{lcccccccclll}
\toprule
\multirow{2}{*}{\textbf{Method}} & \multirow{2}{*}{\textbf{\textit{E}}}  & \multirow{2}{*}{\textbf{\textit{ppl}}} 
& \multicolumn{3}{c}{\textbf{\%Knowledge Acquisition Performance ($\Delta$)}} 
& \multicolumn{3}{c}{\textbf{Entity Learning}} 
& \multicolumn{3}{c}{\textbf{\%Out of Domain Performance ($\Delta$)}} \\

\cmidrule(l{4pt}r{4pt}){4-6} \cmidrule(l{4pt}r{4pt}){7-9} \cmidrule(l{4pt}r{4pt}){10-12}

 &  &  
 & \textbf{Avg.}  & \textbf{High-Freq.} & \textbf{Low-Freq.}  
 & \textbf{\#Acq} & \textbf{\#Ret} & \textbf{\#Forg.}  
 & \textbf{Avg.} & \textbf{MMLU} & \textbf{Social-IQA} \\
\midrule

\multirow{1}{*}{Base} 
& 0 & --  & 41~ & 49~ & 33~ & 17 & -- & -- & 64~ & 50~ & 78~ \\
\midrule

\multirow{5}{*}{\makecell{LoRA}}
& 2 & 8.28 & \metric{45}{41} & \metric{55}{49} & \metric{35}{33} & 3 & 24 & 3 & \metric{54}{64} & \metric{45}{50} & \metric{63}{78} \\
& 4 & 8.05 & \metric{41}{41} & \metric{50}{49} & \metric{32}{33} & 1 & 24 & 3 & \metric{56}{64} & \metric{45}{50} & \metric{68}{78} \\
& 6 & 7.89& \metric{39}{41} & \metric{49}{49} & \metric{30}{33}& 1 & 20 & 7& \metric{56}{64} & \metric{44}{50} & \metric{68}{78} \\
& 8 & 7.76& \metric{38}{41} & \metric{48}{49} & \metric{29}{33}& 0 & 17 & 4& \metric{55}{64} & \metric{44}{50} & \metric{66}{78} \\
& 10 & 7.65& \metric{35}{41} & \metric{43}{49} & \metric{27}{33}& 4 & 9 & 7   & \metric{56}{64} & \metric{43}{50} & \metric{69}{78} \\

\midrule
\multirow{5}{*}{\makecell{Step-wise\\Regular.}}
& 2 & 8.36& \metric{49}{41} & \metric{56}{49} & \metric{42}{33}& 5 & 28 & 0& \metric{60}{64} & \metric{49}{50} & \metric{71}{78} \\
& 4 & 8.08& \metric{48}{41} & \metric{55}{49} & \metric{41}{33}& 3 & 28 & 0& \metric{58}{64} & \metric{48}{50} & \metric{68}{78} \\
& 6 & 7.92& \metric{43}{41} & \metric{52}{49} & \metric{35}{33} & 2 & 26 & 2& \metric{60}{64} & \metric{48}{50} & \metric{72}{78} \\
& 8 & 7.79 & \metric{41}{41} & \metric{49}{49} & \metric{34}{33} & 3 & 19 & 5 & \metric{57}{64} & \metric{46}{50} & \metric{67}{78} \\
& 10 & 7.68& \metric{42}{41} & \metric{50}{49} & \metric{33}{33}& 3 & 16 & 5& \metric{56}{64} & \metric{45}{50} & \metric{68}{78} \\

\midrule
\multirow{5}{*}{\makecell{Pre-trained\\Regular.}}
& 2 & 9.26& \metric{41}{41} & \metric{50}{49} & \metric{32}{33}& 4 & 16 & 2& \metric{63}{64} & \metric{49}{50} & \metric{77}{78} \\
& 4 & 9.23& \metric{42}{41} & \metric{50}{49} & \metric{33}{33}& 2 & 16 & 4& \metric{63}{64} & \metric{50}{50} & \metric{76}{78} \\
& 6 & 9.20& \metric{43}{41} & \metric{51}{49} & \metric{35}{33}& 5 & 14 & 4& \metric{63}{64} & \metric{50}{50} & \metric{77}{78} \\
& 8 & 9.19& \metric{42}{41} & \metric{50}{49} & \metric{33}{33}& 2 & 18 & 2& \metric{63}{64} & \metric{50}{50} & \metric{77}{78} \\
& 10 & 9.16& \metric{43}{41} & \metric{51}{49} & \metric{35}{33}& 4 & 16 & 2& \metric{62}{64} & \metric{50}{50} & \metric{75}{78} \\
 
\midrule
\multirow{5}{*}{\makecell{Structured\\Annotation}}
& 2 & 4.74& \metric{42}{41} & \metric{50}{49} & \metric{34}{33}& 5 & 22 & 3& \metric{53}{64} & \metric{43}{50} & \metric{63}{78} \\
& 4 & 4.63& \metric{43}{41} & \metric{51}{49} & \metric{36}{33}& 4 & 19 & 4& \metric{55}{64} & \metric{44}{50} & \metric{67}{78} \\
& 6 & 4.56& \metric{43}{41} & \metric{51}{49} & \metric{35}{33}& 2 & 22 & 2& \metric{56}{64} & \metric{44}{50} & \metric{67}{78} \\
& 8 & 4.50& \metric{40}{41} & \metric{48}{49} & \metric{32}{33}& 3 & 14 & 6& \metric{54}{64} & \metric{44}{50} & \metric{64}{78} \\
& 10 & 4.45& \metric{37}{41} & \metric{46}{49} & \metric{28}{33}& 4 & 15 & 2& \metric{55}{64} & \metric{44}{50} & \metric{65}{78} \\
 
\midrule

\multirow{5}{*}{\makecell{Curriculum\\Learning}}& 2 & 8.26& \metric{43}{41} & \metric{52}{49} & \metric{35}{33}& 5 & 21 & 0& \metric{57}{64} & \metric{45}{50} & \metric{68}{78} \\
& 4 & 7.97& \metric{43}{41} & \metric{53}{49} & \metric{34}{33}& 4 & 20 & 6& \metric{57}{64} & \metric{46}{50} & \metric{69}{78} \\
& 6 & 7.79& \metric{39}{41} & \metric{48}{49} & \metric{31}{33}& 1 & 15 & 6& \metric{57}{64} & \metric{46}{50} & \metric{67}{78} \\
& 8 & 7.68& \metric{38}{41} & \metric{46}{49} & \metric{31}{33}& 1 & 16 & 2& \metric{56}{64} & \metric{46}{50} & \metric{66}{78} \\
& 10 & 7.61& \metric{36}{41} & \metric{44}{49} & \metric{28}{33}& 1 & 14 & 4& \metric{55}{64} & \metric{45}{50} & \metric{65}{78} \\
 
\bottomrule
\end{tabular}
\end{adjustbox}
\vspace{-2mm}
\caption{Epoch-level learning dynamics for \olmos across different CPT strategies, in terms of knowledge acquisition, and entity learning, as well as out-of-domain task accuracy. The complete table of 10 epochs is presented at \repo.}
\label{tab:olmo_early_epochs_deltaood}
\end{table*}

\begin{table*}[ht]
\small
\centering
\begin{adjustbox}{max width=\textwidth}
\begin{tabular}{lcccccccclll}
\toprule
\multirow{2}{*}{\textbf{Method}} & \multirow{2}{*}{\textbf{\textit{E}}}  & \multirow{2}{*}{\textbf{\textit{ppl}}} 
& \multicolumn{3}{c}{\textbf{\%Knowledge Acquisition Performance ($\Delta$)}} 
& \multicolumn{3}{c}{\textbf{Entity Learning}} 
& \multicolumn{3}{c}{\textbf{\%Out of Domain Performance ($\Delta$)}} \\

\cmidrule(l{4pt}r{4pt}){4-6} \cmidrule(l{4pt}r{4pt}){7-9} \cmidrule(l{4pt}r{4pt}){10-12}

 &  &  
 & \textbf{Avg.}  & \textbf{High-Freq.} & \textbf{Low-Freq.}   & \textbf{\#Acq} & \textbf{\#Ret} & \textbf{\#Forg.}   & \textbf{Avg.} & \textbf{MMLU} & \textbf{Social-IQA} \\
\midrule

\multirow{1}{*}{Base} 
& 0 & --  & 30 & 38 & 22 & -- & -- & -- & 50 & 40 & 59 \\
\midrule

\multirow{5}{*}{\makecell{LoRA}}
& 2 & 11.66 & \metric{35}{30} & \metric{43}{38} & \metric{27}{22} & 3 & 12 & 2 & \metric{36}{50} & \metric{28}{40} & \metric{44}{59} \\
& 4 & 11.06 & \metric{40}{30} & \metric{49}{38} & \metric{32}{22} & 11 & 8 & 5 & \metric{27}{50} & \metric{23}{40} & \metric{31}{59} \\
& 6 & 10.73 & \metric{40}{30} & \metric{48}{38} & \metric{31}{22} & 3 & 14 & 5 & \metric{31}{50} & \metric{27}{40} & \metric{36}{59} \\
& 8 & 10.49 & \metric{40}{30} & \metric{48}{38} & \metric{31}{22} & 3 & 17 & 6 & \metric{28}{50} & \metric{25}{40} & \metric{31}{59} \\
& 10 & 10.30 & \metric{40}{30} & \metric{48}{38} & \metric{31}{22} & 1 & 16 & 5 & \metric{26}{50} & \metric{24}{40} & \metric{27}{59} \\

\midrule
\multirow{5}{*}{\makecell{Step-wise\\Regular.}}
& 2 & 11.84 & \metric{35}{30} & \metric{43}{38} & \metric{26}{22} & 6 & 8 & 2 & \metric{45}{50} & \metric{38}{40} & \metric{52}{59} \\
& 4 & 11.16 & \metric{35}{30} & \metric{43}{38} & \metric{28}{22} & 0 & 10 & 2 & \metric{35}{50} & \metric{32}{40} & \metric{37}{59} \\
& 6 & 10.82 & \metric{37}{30} & \metric{45}{38} & \metric{29}{22} & 4 & 9 & 0 & \metric{28}{50} & \metric{31}{40} & \metric{25}{59} \\
& 8 & 10.58 & \metric{38}{30} & \metric{45}{38} & \metric{30}{22} & 1 & 13 & 2 & \metric{29}{50} & \metric{32}{40} & \metric{27}{59} \\
& 10 & 10.40 & \metric{36}{30} & \metric{44}{38} & \metric{28}{22} & 3 & 12 & 3 & \metric{35}{50} & \metric{34}{40} & \metric{36}{59} \\

\midrule
\multirow{5}{*}{\makecell{Pre-training\\Regular.}}
& 2 & 13.53 & \metric{31}{30} & \metric{39}{38} & \metric{22}{22} & 0 & 11 & 5 & \metric{48}{50} & \metric{40}{40} & \metric{56}{59} \\
& 4 & 13.42 & \metric{31}{30} & \metric{39}{38} & \metric{22}{22} & 2 & 9 & 1 & \metric{48}{50} & \metric{42}{40} & \metric{55}{59} \\
& 6 & 13.38 & \metric{31}{30} & \metric{39}{38} & \metric{23}{22} & 0 & 11 & 0 & \metric{49}{50} & \metric{41}{40} & \metric{57}{59} \\
& 8 & 13.35 & \metric{31}{30} & \metric{39}{38} & \metric{23}{22} & 1 & 11 & 0 & \metric{49}{50} & \metric{41}{40} & \metric{56}{59} \\
& 10 & 13.33 & \metric{32}{30} & \metric{39}{38} & \metric{24}{22} & 2 & 9 & 1 & \metric{49}{50} & \metric{41}{40} & \metric{56}{59} \\

\midrule
\multirow{5}{*}{\makecell{Structured\\Annotation}}
& 2 & 6.30 & \metric{33}{30} & \metric{42}{38} & \metric{23}{22} & 5 & 9 & 1 & \metric{33}{50} & \metric{28}{40} & \metric{37}{59} \\
& 4 & 6.05 & \metric{36}{30} & \metric{45}{38} & \metric{26}{22} & 2 & 10 & 6 & \metric{30}{50} & \metric{27}{40} & \metric{33}{59} \\
& 6 & 5.91 & \metric{34}{30} & \metric{43}{38} & \metric{24}{22} & 5 & 6 & 6 & \metric{21}{50} & \metric{21}{40} & \metric{21}{59} \\
& 8 & 5.82 & \metric{36}{30} & \metric{45}{38} & \metric{27}{22} & 7 & 7 & 2 & \metric{30}{50} & \metric{23}{40} & \metric{37}{59} \\
& 10 & 5.74 & \metric{33}{30} & \metric{42}{38} & \metric{23}{22} & 2 & 7 & 5 & \metric{26}{50} & \metric{22}{40} & \metric{30}{59} \\

\midrule
\multirow{5}{*}{\makecell{Curriculum\\Learning}}
& 2 & 11.62 & \metric{35}{30} & \metric{44}{38} & \metric{27}{22} & 3 & 10 & 6 & \metric{39}{50} & \metric{32}{40} & \metric{46}{59} \\
& 4 & 10.94 & \metric{38}{30} & \metric{47}{38} & \metric{30}{22} & 5 & 10 & 4 & \metric{28}{50} & \metric{28}{40} & \metric{27}{59} \\
& 6 & 10.64 & \metric{38}{30} & \metric{45}{38} & \metric{30}{22} & 3 & 13 & 5 & \metric{31}{50} & \metric{29}{40} & \metric{34}{59} \\
& 8 & 10.52 & \metric{39}{30} & \metric{46}{38} & \metric{31}{22} & 5 & 7 & 5 & \metric{26}{50} & \metric{28}{40} & \metric{24}{59} \\
& 10 & 10.13 & \metric{39}{30} & \metric{47}{38} & \metric{30}{22} & 5 & 13 & 8 & \metric{31}{50} & \metric{25}{40} & \metric{37}{59} \\
\bottomrule
\end{tabular}
\end{adjustbox}
\vspace{-2mm}
\caption{Epoch-level learning dynamics for \llama\textsubscript{1B} across different CPT strategies, in terms of knowledge acquisition, and entity learning, as well as out-of-domain task accuracy. The complete table of 10 epochs is presented at \repo.}
\label{tab:llama1b_early_epochs_deltaood}
\end{table*}

\begin{table*}[ht]
\centering
\small
\begin{adjustbox}{max width=\textwidth}
\begin{tabular}{lcccccccclll}
\toprule
\multirow{2}{*}{\textbf{Method}} & \multirow{2}{*}{\textbf{\textit{E}}}  & \multirow{2}{*}{\textbf{\textit{ppl}}} 
& \multicolumn{3}{c}{\textbf{\%Knowledge Acquisition Performance ($\Delta$)}} 
& \multicolumn{3}{c}{\textbf{Entity Learning}} 
& \multicolumn{3}{c}{\textbf{\%Out of Domain Performance ($\Delta$)}} \\

\cmidrule(l{4pt}r{4pt}){4-6} \cmidrule(l{4pt}r{4pt}){7-9} \cmidrule(l{4pt}r{4pt}){10-12}

 &  &  
 & \textbf{Avg.}  & \textbf{High-Freq.} & \textbf{Low-Freq.}  
 & \textbf{\#Acq} & \textbf{\#Ret} & \textbf{\#Forg.}  
 & \textbf{Avg.} & \textbf{MMLU} & \textbf{Social-IQA} \\
\midrule

\multirow{1}{*}{Base} 
& 0 & -- & 52~ & 60~ & 44~ &  & -- & -- & 72~ & 66~ & 78~ \\
\midrule

\multirow{5}{*}{LoRA}
& 2 & 7.16   & \metric{52}{52} & \metric{61}{60} & \metric{44}{44}   & 5 & 35 & 5   & \metric{54}{72} & \metric{47}{66} & \metric{61}{78} \\
& 4 & 6.54   & \metric{48}{52} & \metric{58}{60} & \metric{39}{44}   & 1 & 28 & 10   & \metric{53}{72} & \metric{50}{66} & \metric{56}{78} \\
& 6 & 6.12   & \metric{51}{52} & \metric{60}{60} & \metric{42}{44}   & 6 & 29 & 8   & \metric{50}{72} & \metric{46}{66} & \metric{54}{78} \\
& 8 & 5.80   & \metric{51}{52} & \metric{59}{60} & \metric{42}{44}   & 4 & 28 & 7   & \metric{46}{72} & \metric{46}{66} & \metric{46}{78} \\
& 10 & 5.56   & \metric{51}{52} & \metric{60}{60} & \metric{42}{44}   & 6 & 27 & 9   & \metric{44}{72} & \metric{38}{66} & \metric{51}{78} \\

\midrule
\multirow{5}{*}{\makecell{Step-wise\\Regular.}}
& 2 & 7.25   & \metric{52}{52} & \metric{62}{60} & \metric{43}{44}   & 1 & 34 & 5   & \metric{59}{72} & \metric{52}{66} & \metric{67}{78} \\
& 4 & 6.65   & \metric{53}{52} & \metric{61}{60} & \metric{45}{44}   & 5 & 35 & 5   & \metric{56}{72} & \metric{56}{66} & \metric{57}{78} \\
& 6 & 6.27   & \metric{52}{52} & \metric{63}{60} & \metric{42}{44}   & 3 & 35 & 4   & \metric{55}{72} & \metric{52}{66} & \metric{57}{78} \\
& 8 & 5.99   & \metric{52}{52} & \metric{60}{60} & \metric{44}{44}   & 12 & 32 & 3   & \metric{51}{72} & \metric{48}{66} & \metric{54}{78} \\
& 10 & 5.74   & \metric{54}{52} & \metric{62}{60} & \metric{46}{44}   & 8 & 36 & 7   & \metric{50}{72} & \metric{50}{66} & \metric{50}{78} \\

\midrule
\multirow{5}{*}{\makecell{Pre-training\\Regular.}}
& 2 & 8.05   & \metric{52}{52} & \metric{61}{60} & \metric{43}{44}   & 6 & 31 & 4   & \metric{69}{72} & \metric{63}{66} & \metric{74}{78} \\
& 4 & 7.95   & \metric{52}{52} & \metric{60}{60} & \metric{44}{44}   & 2 & 33 & 5   & \metric{71}{72} & \metric{64}{66} & \metric{77}{78} \\
& 6 & 7.90   & \metric{53}{52} & \metric{62}{60} & \metric{44}{44}   & 2 & 35 & 4   & \metric{69}{72} & \metric{64}{66} & \metric{74}{78} \\
& 8 & 7.87   & \metric{53}{52} & \metric{61}{60} & \metric{44}{44}   & 4 & 33 & 4   & \metric{68}{72} & \metric{64}{66} & \metric{72}{78} \\
& 10 & 7.85   & \metric{53}{52} & \metric{61}{60} & \metric{45}{44}   & 5 & 34 & 2   & \metric{69}{72} & \metric{64}{66} & \metric{75}{78} \\

\midrule
\multirow{5}{*}{\makecell{Structured\\Annotation}}
& 2 & 4.38   & \metric{52}{52} & \metric{61}{60} & \metric{44}{44}   & 6 & 32 & 5   & \metric{58}{72} & \metric{54}{66} & \metric{62}{78} \\
& 4 & 4.09   & \metric{50}{52} & \metric{60}{60} & \metric{41}{44}   & 2 & 32 & 5   & \metric{51}{72} & \metric{54}{66} & \metric{48}{78} \\
& 6 & 3.89   & \metric{47}{52} & \metric{57}{60} & \metric{38}{44}   & 3 & 29 & 5   & \metric{54}{72} & \metric{50}{66} & \metric{57}{78} \\
& 8 & 3.75   & \metric{49}{52} & \metric{58}{60} & \metric{41}{44}   & 9 & 20 & 4   & \metric{45}{72} & \metric{43}{66} & \metric{47}{78} \\
& 10 & 3.62   & \metric{45}{52} & \metric{54}{60} & \metric{37}{44}   & 6 & 19 & 12   & \metric{43}{72} & \metric{38}{66} & \metric{48}{78} \\

\midrule
\multirow{5}{*}{\makecell{Curriculum\\Learning}}
& 2 & 7.22   & \metric{53}{52} & \metric{62}{60} & \metric{44}{44}   & 8 & 31 & 2   & \metric{50}{72} & \metric{54}{66} & \metric{46}{78} \\
& 4 & 6.36   & \metric{47}{52} & \metric{57}{60} & \metric{38}{44}   & 1 & 29 & 6   & \metric{54}{72} & \metric{52}{66} & \metric{55}{78} \\
& 6 & 5.93   & \metric{49}{52} & \metric{58}{60} & \metric{39}{44}   & 4 & 30 & 3   & \metric{45}{72} & \metric{44}{66} & \metric{46}{78} \\
& 8 & 5.66   & \metric{48}{52} & \metric{56}{60} & \metric{39}{44}   & 5 & 25 & 6   & \metric{48}{72} & \metric{45}{66} & \metric{50}{78} \\
& 10 & 5.38   & \metric{47}{52} & \metric{56}{60} & \metric{38}{44}   & 8 & 22 & 7   & \metric{57}{72} & \metric{51}{66} & \metric{63}{78} \\
\bottomrule
\end{tabular}
\end{adjustbox}
\vspace{-2mm}
\caption{Epoch-level learning dynamics for \llama\textsubscript{8B} across different CPT strategies, in terms of knowledge acquisition, and entity learning, as well as out-of-domain task accuracy. The complete table of 10 epochs is presented at \repo.}
\label{tab:llama8b_early_epochs_deltaood}
\end{table*}

\begin{figure*}[!ht]
    \centering
    \includegraphics[width=0.92\linewidth]{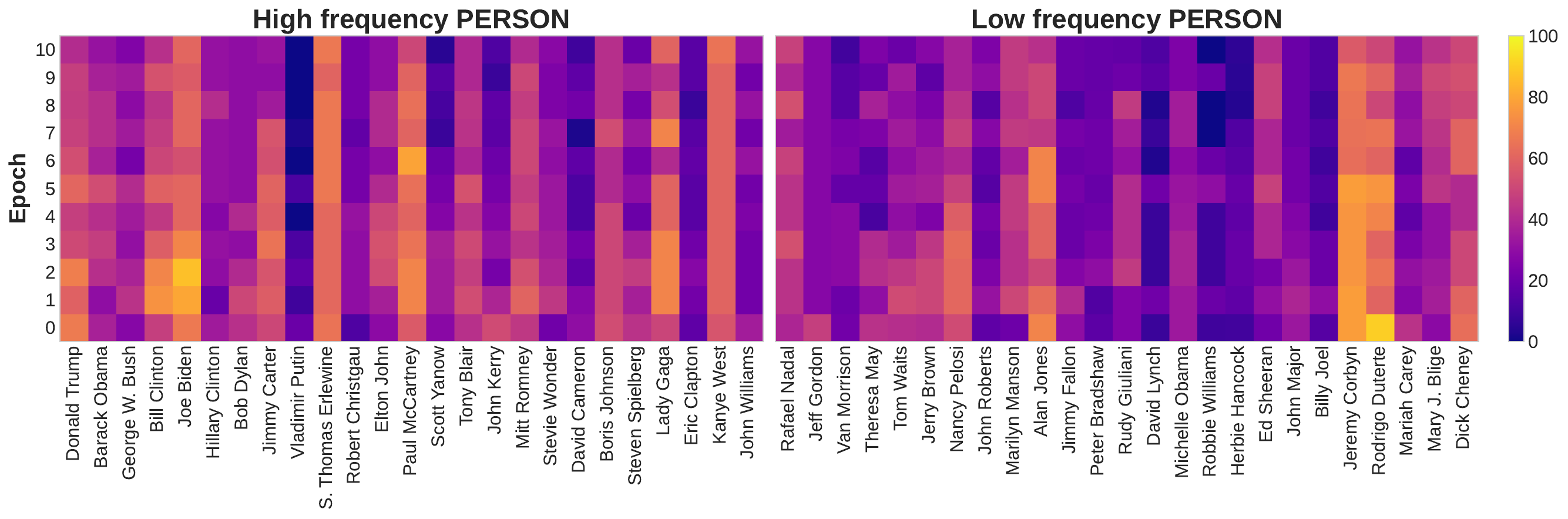}
\caption{\textbf{Learning dynamics for high- and low-frequency named entities (PERSON) via LoRA across continual pre-training epoch (\olmo)}. Generally Low-frequency entities display minimal learning throughout, while high-frequency entities exhibit non-monotonic improvements across epochs.}
    \label{fig:olmo_lora_freq_heat_person}
\end{figure*}

\begin{figure*}[!ht]
    \centering
    \includegraphics[width=0.92\linewidth]{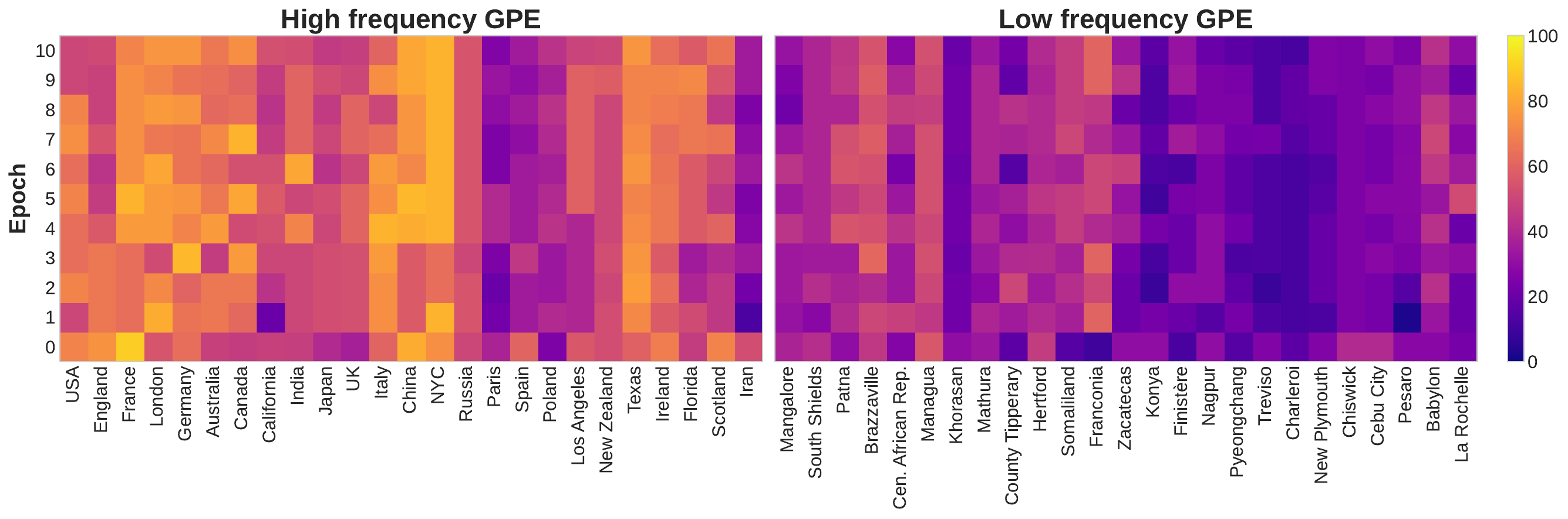}
\caption{\textbf{Learning dynamics for high- and low-frequency named entities (GPE) via LoRA across continual pre-training epoch (\llama\textsubscript{1B})}.}
    \label{fig:llama3-1b_lora_freq_heat_gpe}
\end{figure*}

\begin{figure*}[!ht]
    \centering
    \includegraphics[width=0.92\linewidth]{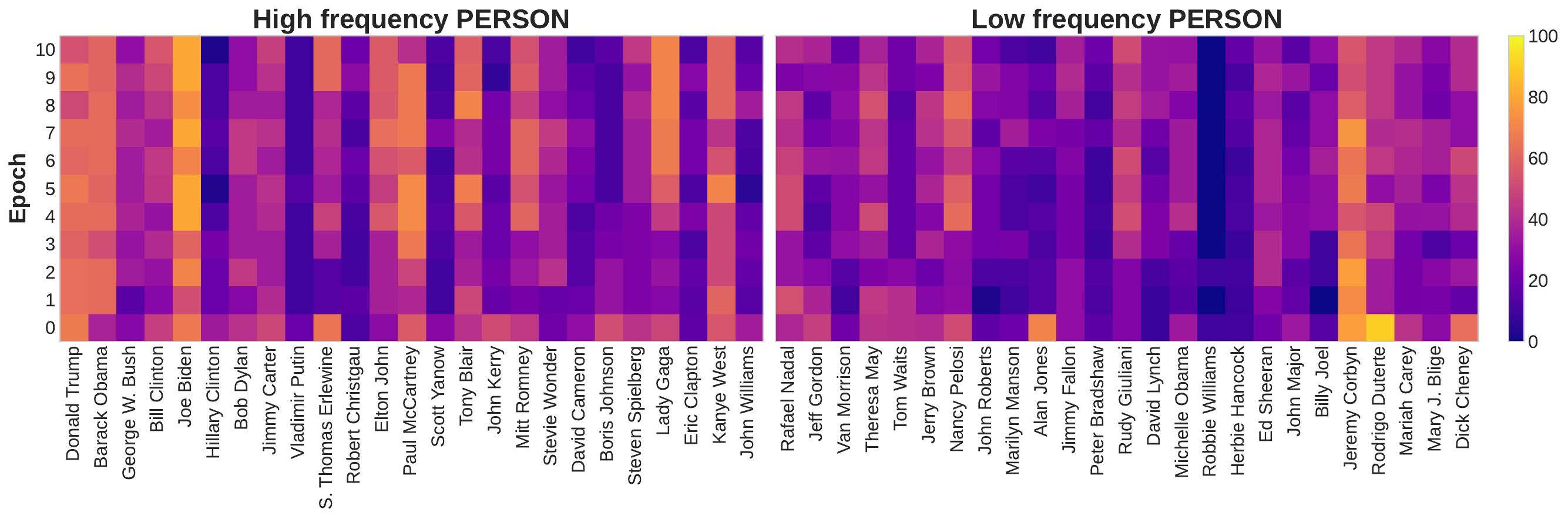}
\caption{\textbf{Learning dynamics for high- and low-frequency named entities (PERSON) via LoRA across continual pre-training epoch (\llama\textsubscript{1B})}.}
    \label{fig:llama3-1b_lora_freq_heat_person}
\end{figure*}

\begin{figure*}[!ht]
    \centering
    \includegraphics[width=0.92\linewidth]{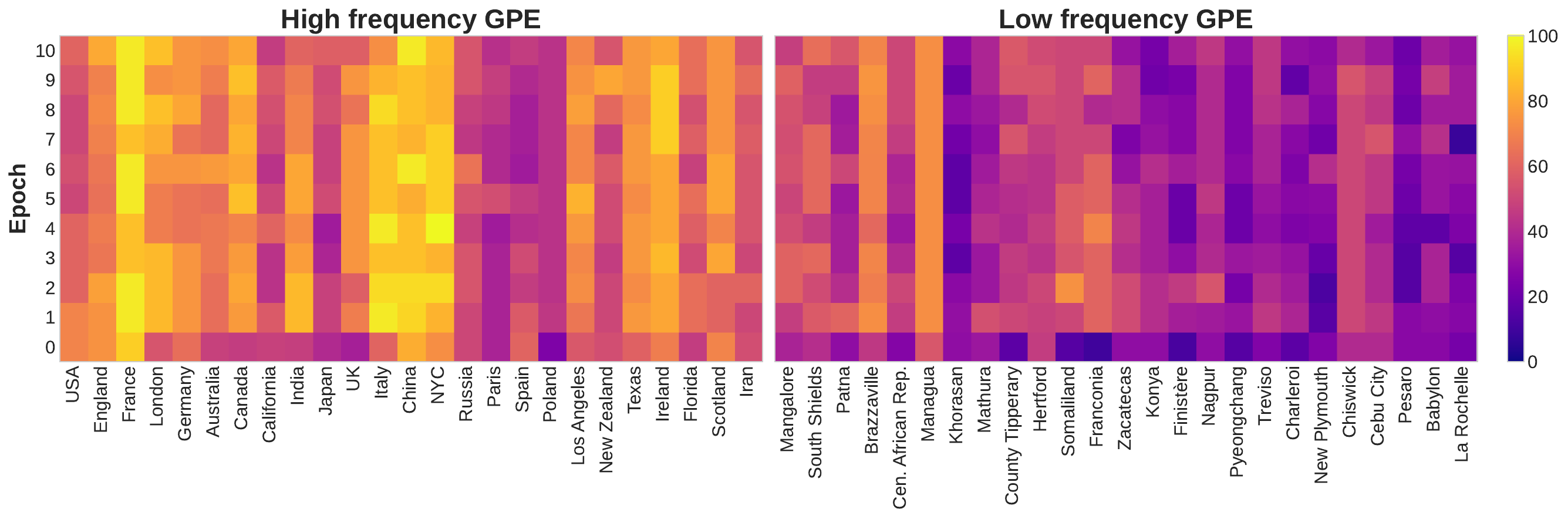}
\caption{\textbf{Learning dynamics for high- and low-frequency named entities (GPE) via LoRA across continual pre-training epoch (\llama\textsubscript{8B})}.}
    \label{fig:llama3-8b_lora_freq_heat_gpe}
\end{figure*}

\begin{figure*}[ht!]
    \centering
    \includegraphics[width=0.92\linewidth]{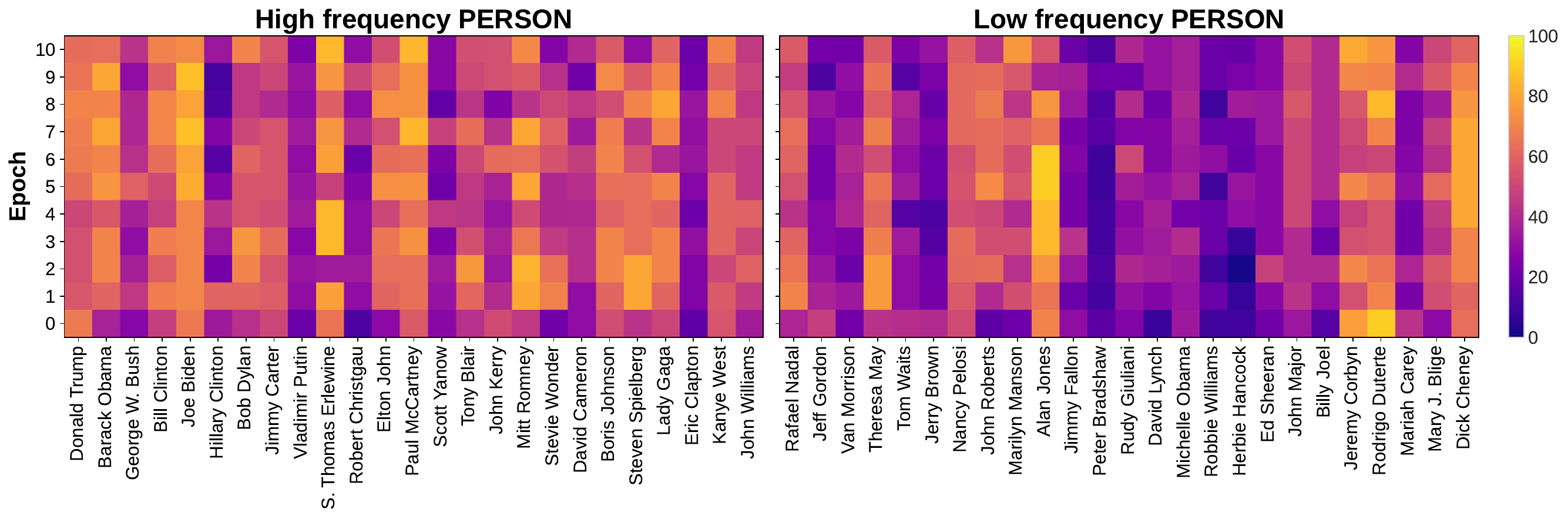}
\caption{\textbf{Learning dynamics for high- and low-frequency named entities (PERSON) via LoRA across continual pre-training epoch (\llama\textsubscript{8B})}.}
    \label{fig:llama3-8b_lora_freq_heat_person}
\end{figure*}

\begin{table*}[t]

\centering
\setlength{\tabcolsep}{4pt}
\renewcommand{\arraystretch}{1.15}
\begin{adjustbox}{max width=\linewidth}
\begin{tabular}{>{\centering\arraybackslash}m{2cm}ccc}
\toprule
\makecell{\textbf{Frequency}} & {\textbf{GPE}}& {\textbf{PERSON}}\\
\cmidrule(l{4pt}r{4pt}){1-1} \cmidrule(l{4pt}r{4pt}){2-2} \cmidrule(l{4pt}r{4pt}){3-3}
\multirow{5}{*}{\textbf{High}} &
\textit{USA~,~England~,~France~,~London~,~Germany} &\textit{Donald Trump~,~Barack Obama~,~George W.\ Bush~,~Bill Clinton~,~Joe Biden} \\
& \textit{Australia~,~Canada~,~California~,~India~,~Japan} & \textit{Hillary Clinton~,~Bob Dylan~,~Jimmy Carter~,~Vladimir Putin~,~Stephen T.\ Erlewine} \\
 & \textit{UK~,~Italy~,~China~,~New Zealand,~Russia} & \textit{Robert Christgau~,~Elton John~,~Paul McCartney~,~Scott Yanow~,~Tony Blair} \\

& \textit{Paris~,~Spain~,~Poland~,~Los Angeles~,~NYC~} & \textit{John Kerry~,~Mitt Romney~,~Stevie Wonder~,~David Cameron~,~Boris Johnson} \\

& \textit{Texas~,~Ireland~,~Florida~,~Scotland~,~Iran} & \textit{Steven Spielberg~,~Lady Gaga~,~Eric Clapton~,~Kanye West~,~John Williams} \\

\midrule
\multirow{5}{*}{\textbf{Low}} &
\textit{Mangalore~,~South Shields~,~Patna~,~Brazzaville~,~Finist\`ere}  &
\textit{Rafael Nadal~,~Jeff Gordon~,~Van Morrison~,~Theresa May~,~Tom Waits} \\
& \textit{Managura~,~Khorasan~,~Mathura~,~County Tipperary~,~Hertford} & \textit{Jerry Brown~,~Nancy Pelosi~,~John Roberts~,~Marilyn Manson~,~Alan Jones}\\
& \textit{Somaliland~,~Franconia~,~Zacatecas~,~Konya~,~Cen. African Rep.}& \textit{Jimmy Fallon~,~Peter Bradshaw~,~Rudy Giuliani~,~David Lynch~,~Michelle Obama} \\
& \textit{Nagpur~,~Pyeongchang~,~Treviso~,~Charleroi~,~New Plymouth}& \textit{Robbie Williams~,~Herbie Hancock~,~Ed Sheeran~,~John Major~,~Billy Joel} \\
& \textit{Chiswick~,~Cebu City~,~Pesaro~,~Babylon~,~La Rochelle}& \textit{Jeremy Corbyn~,~Rodrigo Duterte~,~Mariah Carey~,~Mary J.\ Blige~,~Dick Cheney} \\
\bottomrule
\end{tabular}
\end{adjustbox}
\caption{High- and low-frequency GPE and PERSON entities.}
\label{tab:gpe-entities}
\end{table*}

\end{document}